%% file: main.tex
\documentclass[letterpaper, 10 pt, journal, twoside]{IEEEtran}

\usepackage{times}
\usepackage{graphics}
\usepackage{graphicx}
\usepackage{amsmath,amssymb,amsopn,amstext,amsfonts}
\usepackage[ruled,vlined]{algorithm2e}
\SetKwComment{Comment}{$\triangleright$\ }{}
\usepackage{cancel}
\usepackage[space]{cite}
\usepackage{pdfsync}
\usepackage{balance}
\usepackage{color}
\usepackage{mathtools}
\usepackage{bm}

\usepackage{diagbox}
\usepackage{float}
\usepackage{epstopdf}
\usepackage{pifont}
\usepackage{fixltx2e}
\usepackage{amsmath}
\usepackage{multirow}
\usepackage{url}
\usepackage{verbatim}
\usepackage{caption}
\usepackage{adjustbox}
\usepackage{booktabs}
\usepackage{threeparttable}
\usepackage{makecell}
\usepackage{pbox}
\usepackage{subfig}
\makeatletter
\let\NAT@parse\undefined
\makeatother
\usepackage[linkcolor=red,citecolor=green,urlcolor=red,colorlinks=true,urlcolor=black]{hyperref}
\usepackage{xcolor}
\usepackage{pifont}
\newcommand{\cmark}{\color{green}\ding{52}}

\newcommand{\xmark}{\color{red}\ding{54}}

\usepackage{ulem}

\usepackage{siunitx}
\usepackage{tikz}

\usepackage{hhline}
\usepackage{ulem}
\usepackage{dblfloatfix}
\usepackage{placeins}
\newcommand{\novalue}{{\textendash}}
\usepackage{siunitx}

\input{chapters/math_macros.tex}

\begin{document}
\title{EvTTC: An Event Camera Dataset for Time-to-Collision Estimation}

\author
{
	Kaizhen Sun$^{1, \ast}$, Jinghang Li$^{1, \ast}$, Kuan Dai$^{1}$, Bangyan Liao$^{2}$, Wei Xiong$^{3}$ and Yi Zhou$^{1, \dagger}$
	\thanks{Manuscript received: December 5, 2024; Revised: February, 21, 2025; Accepted: April, 17, 2025.}
	\thanks{This paper was recommended for publication by Editor Aniket Bera upon evaluation of the Associate Editor and Reviewers' comments. 
		This work was supported by the National Key Research and Development Project of China under Grant 2023YFB4706600.}
	\thanks{\textsuperscript{1}Neuromorphic Automation and Intelligence Lab (NAIL), School of Robotics, Hunan University, Changsha, China.}
	\thanks{\textsuperscript{2}School of Engineering, Westlake University, China.}
	\thanks{\textsuperscript{3}Xidi Zhijia (Hunan) Co., Ltd., Changsha, China.}
        \thanks{$\ast$ Authors contributed equally.}
        \thanks{$\dagger$ Corresponding author: Yi Zhou. Email: {\tt\small eeyzhou@hnu.edu.cn}.}
	\thanks{Digital Object Identifier (DOI): see top of this page.}
}

\markboth{IEEE Robotics and Automation Letters. Preprint Version. Accepted April, 2025}
{Sun \MakeLowercase{\textit{et al.}}: EvTTC}

\setcounter{figure}{-2} %
\makeatletter
\g@addto@macro\@maketitle{
\vspace{4ex}
\input{floats/fig_eye_catcher}
\vspace{-4ex}
}
\makeatother
\maketitle

\vspace*{2pt}
\input{chapters/00_Abstract}


\input{chapters/Multimedia_material}
\input{chapters/01_Introduction}
\input{chapters/02_Related_Work}
\input{chapters/03_Hardware_Setup}

\input{chapters/04_Sequences_Overview}
\input{chapters/Collision_Simulation_SLIDER}

\input{chapters/05_Experiments}
\input{chapters/06_Conclusion}

\bibliographystyle{IEEEtran}
\normalem
\bibliography{myBib}
\input{chapters/Appendix}

\end{document}

%% file: chapters/math_macros.tex
\newcommand{\etal}{\textit{et al}.}

\newcommand{\eg}{\textit{e}.\textit{g}.}

\newcommand*\circled[1]{\tikz[baseline=(char.base)]{\node[shape=circle,draw,inner sep=1pt] (char) {#1};}}

%% file: floats/fig_eye_catcher.tex
\centering
\includegraphics[width=0.98\linewidth]{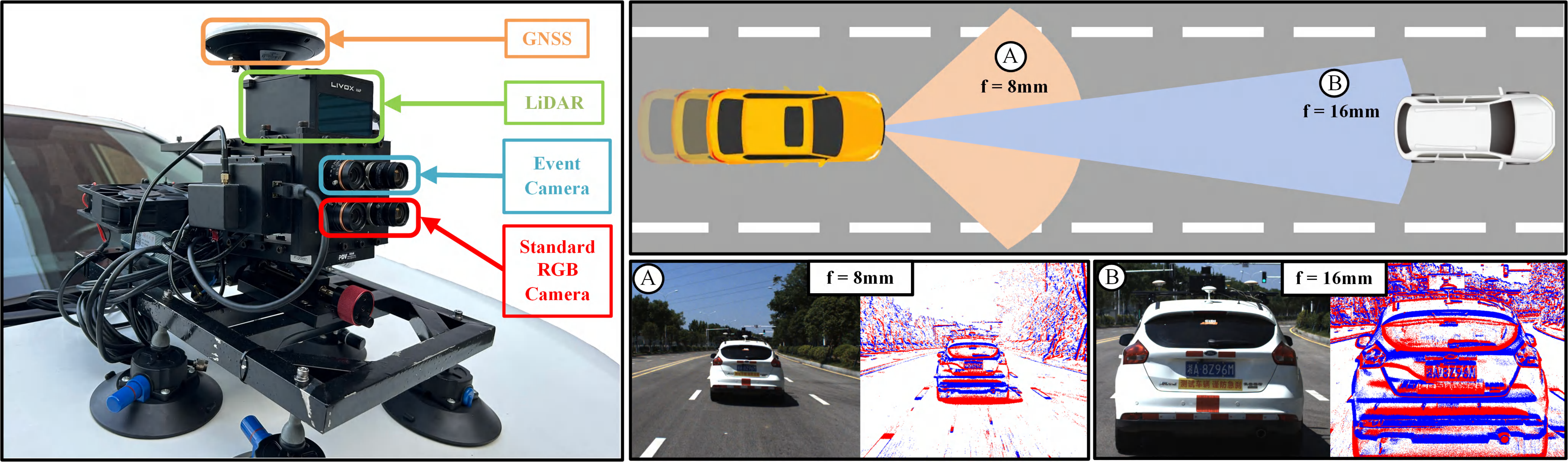}
\captionof{figure}{
Left: An overview of the data collection setup. Right: The top image shows a bird's-eye view of the vehicle on the highway, with fields of view and detection distances indicated by circled letters. The bottom row presents RGB images and accumulated event data captured by both the 8-mm Lens and 16-mm Lens
Camera pairs.
}
\label{fig:eye_catcher}

%% file: chapters/00_Abstract.tex
\begin{abstract}
Time-to-Collision (TTC) estimation lies in the core of the forward collision warning (FCW) functionality, which is key to all Automatic Emergency Braking (AEB) systems.
Although the success of solutions using frame-based cameras (\eg, \textit{Mobileye}'s solutions) has been witnessed in normal situations, some extreme cases, such as the sudden variation in the relative speed of leading vehicles and the sudden appearance of pedestrians, still pose significant risks that cannot be handled.
This is due to the inherent imaging principles of frame-based cameras, where the time interval between adjacent exposures introduces considerable system latency to AEB.
Event cameras, as a novel bio-inspired sensor, offer ultra-high temporal resolution and can asynchronously report brightness changes at the microsecond level.
To explore the potential of event cameras in the above-mentioned challenging cases, we propose EvTTC, which is, to the best of our knowledge, the first multi-sensor dataset focusing on TTC tasks under high-relative-speed scenarios. 
EvTTC consists of data collected using standard cameras and event cameras, covering various potential collision scenarios in daily driving and involving multiple collision objects. 
Additionally, LiDAR and GNSS/INS measurements are provided for the calculation of ground-truth TTC.
Considering the high cost of testing TTC algorithms on full-scale mobile platforms, we also provide a small-scale TTC testbed for experimental validation and data augmentation. 
All the data and the design of the testbed are open sourced, and they can serve as a benchmark that will facilitate the development of vision-based TTC techniques.
\end{abstract}

%% file: chapters/Multimedia_material.tex
\section*{Multimedia material}
\label{sec:multimedia material}

Supplemental video: \href{https://youtu.be/qJC016GK888}{\textcolor{red}{https://youtu.be/qJC016GK888}}

EvTTC Benchmark: \href{https://nail-hnu.github.io/EvTTC/}{\textcolor{red}{https://nail-hnu.github.io/EvTTC/}}

%% file: chapters/01_Introduction.tex
\section{Introduction}
\label{sec: introduction}
\input{floats/tab_dataset_compare}
Time-to-Collision (TTC) refers to the time it takes for two objects to collide under their current speed. 
Automated Emergency Braking (AEB) systems, which typically include the Forward Collision Warning (FCW) functionality, are now standard features in automobiles.
They apply the vehicle’s foundation brakes to prevent a front crash if the driver does not intervene. 
The key component of FCW is estimating the TTC, which refers to the time it takes for a potential collision under current speed.
A large number of technical reports on AEB and FCW have demonstrated that these functionalities reduce front-to-rear crashes by $27\%$, rear-end collisions by $27\%–50\%$, and rear-end injury crashes by $35\%–56\%$~\cite{cicchino2017effectiveness},~\cite{fildes2015effectiveness},~\cite{isaksson2015evaluation},~\cite{rizzi2014injury}.

Compared to LiDAR or radar-based solutions ~\cite{gawande2022autonomous},~\cite{bosnak2017efficient},~\cite{kotur2021camera}, ~\cite{venkatesha2023detection}, traditional frame-based cameras are a more popular choice due to their low cost.
There is extensive research on TTC estimation using traditional cameras~\cite{dagan2004forward},~\cite{meyer1992estimation},~\cite{negre2008real},~\cite{stabinger2016monocular}.
The most common method calculates TTC by analyzing two consecutive images of a monocular camera.
The scale of the leading vehicle in the image changes due to its relative motion to the host vehicle, and this variation of scale is used to estimate the TTC.
However, this approach is constrained by the frame rate of traditional cameras, which typically operate at 10 Hz in ADAS to balance cost, bandwidth, and power consumption. 
The resulting 100 ms interval between frames, even before accounting for computation time, can lead to significant latency in FCW and AEB, especially in high-speed scenarios.

Event-based cameras, inspired by the biological mechanisms of the human visual system, operate differently from traditional frame-based cameras. 
Rather than capturing entire frames, event-based cameras asynchronously report changes in brightness at individual pixels. 
This approach offers spatio-temporal sparsity, microsecond-level temporal resolution, and high dynamic range, making event-based cameras well-suited for perception tasks involving rapid motion, such as robotics and autonomous driving~\cite{falanga2020dynamic},~\cite{rebecq2017real},~\cite{lu2023event},~\cite{gehrig2024low}.

Despite the potential of event-based cameras, research in the field of TTC estimation faces significant challenges, particularly due to the lack of suitable datasets focusing on high relative-speed scenarios.
To fill this gap, we present a new dataset (see Fig.~\ref{fig:eye_catcher}) that can serve as a benchmark featuring collision scenarios with high relative speed, multiple collision targets, and diverse scenes, facilitating the development and evaluation of TTC estimation methods.
The benefits brought by the proposed dataset consist of:
\begin{itemize} 
    \item A diverse set of sequences featuring various targets, such as real vehicles, inflatable vehicles, and dummies, across a wide range of relative speeds, including both routine and challenging situations. 
    \item A low-cost and small-scale TTC testbed that facilitates the generation of quasi-real data at different relative speeds.
    The design of the testbed is open-source.
    \item A specific benchmark for the TTC task that can serve as an evaluation platform for the community to test and compare different TTC estimation methods.
\end{itemize}

%% file: floats/tab_dataset_compare.tex
\begin{table*}[]
    \centering

    \renewcommand{\arraystretch}{1.2}
    \begin{tabular}{lcccccc}
    \toprule
    \multirow{2}{*}{\textbf{Dataset}} & \multicolumn{2}{l|}{\textbf{Frame Camera}}                                           & \multicolumn{2}{l}{\textbf{Event Camera}}                       & \multicolumn{1}{c}{\textbf{Emergency}}  & \multirow{2}{*}{\textbf{Groundtruth}} \\ \cline{2-5} 
                                      & \multicolumn{1}{c}{Resolution {[}pix{]}}    & \multicolumn{1}{l|}{Detection Range {[}m{]}} & Resolution {[}pix{]}   & \multicolumn{1}{l}{Detection Range {[}m{]}}   & \multicolumn{1}{c}{\textbf{Break}}   &                                       \\ \midrule
    {DR~(eye)~VE}~\cite{palazzi2018predicting}   & 1920$\times$1080                         & \novalue                                     & \novalue                  & \novalue                                      & \xmark                               &  {Driver Attention Map}                        \\
    {DADA-2000}~\cite{fang2019dada}     & 1584$\times$660                                        & \novalue                                     & \novalue                  & \novalue                                      & \cmark                               &  {Driver Attention Map}                          \\
    {Crash to Not Crash}~\cite{kim2019crash}         & 710$\times$400                                        & \novalue                           & \novalue                  & \novalue                                   & \cmark                               &  {2D Bounding Box}                            \\ 
    {TSTTC}~\cite{shi2023tsttc}         & 1024$\times$576                                        & {[}16{-}96{]}                           & \novalue                  & \novalue                                      & \xmark                               &  \textbf{TTC}                           \\ 
    MVSEC~\cite{zhu2018multivehicle}  & 752$\times$480                                        & 28                                      & 346$\times$260                  & 12                                       & \xmark                               & {Depth, GPS}                            \\ 
    {DSEC}~\cite{Gehrig21ral}           & 1440$\times$1080                                        & 80                                      & 640$\times$480                  & 35                                       & \xmark                               & {Depth, RTK GPS}                            \\ 
    % VECTOR~\cite{gao2022vector}       & 1.3                                     & 58                                         & 0.3                  & 34                                       & \xmark                               & Depth, GPS                            \\
    ViViD++~\cite{lee2022vivid++}     & 1280$\times$1024                                        & 53                                      & 640$\times$480                  & 50                                       & \xmark                               & {Depth, RTK GPS}                            \\
    M3ED~\cite{chaney2023m3ed}        & 1280$\times$800                & 68                                      & \textbf{1280$\times$720}         & 65                                       & \xmark                               & {Depth, RTK GPS}                            \\
    
    \midrule
    \textbf{EvTTC} (Ours)              & \textbf{1920$\times$1200}                               & \textbf{{[}160{-}295{]}}                & \textbf{1280$\times$720}         & \textbf{{[}99{-}197{]}}                  & \cmark                               & \textbf{TTC}{, Depth, RTK GPS}              \\ \bottomrule
    \end{tabular} 
    
    \caption{\textit{Comparison of different event camera datasets in driving scenarios.} The symbol "\novalue" indicates \emph{not available}. Emergency braking represents whether there is a scenario with a rapid decrease in vehicle speed and a small TTC value.}
    \label{tab:datasets_compare}
\end{table*}

%% file: chapters/02_Related_Work.tex
\section{Related Work}
\label{sec: related work}

In this section, our literature review focuses on three aspects: 
1) event-based TTC estimation methods,
2) autonomous driving datasets focused on collision risk prediction and TTC estimation tasks, and 3) event camera-centric datasets.
Table~\ref{tab:datasets_compare} provides an overview of these datasets, briefly showcasing the different sensor configurations they use and their different focuses.

\textbf{1) \textit{Event-based TTC Estimation Methods:}} Inspired by methods using standard cameras~\cite{meyer1992estimation},~\cite{stabinger2016monocular}, most event-based approaches~\cite{clady2014asynchronous},~\cite{dupeyroux2021neuromorphic},~\cite{dinaux2021faith} are built on optical flow derived from event data. 
Among these methods, \cite{clady2014asynchronous} and~\cite{dinaux2021faith} typically use the estimated optical flow to select the Focus of Expansion (FoE) located in the negative half-plane for TTC estimation. 
However, the performance of these methods is significantly influenced by the accuracy of the estimated FoE.
Gallego \etal ~\cite{gallego2018unifying} propose the CMax, which can implicitly handle the data association between events.
This approach enables precise estimation of TTC through geometric model fitting (\textit{e.g.}, an affine model).
To address the event collapse issue in CMax, Shiba \etal ~\cite{shiba2023fast} present a regularization method based on the CMax framework.
ETTCM~\cite{nunes2023time} extends the DMin framework~\cite{nunes2021robust} to jointly estimate the scene’s inverse depth and global motion to generate the TTC map.
Further advancing the field, Li \etal ~\cite{li2024eventaidedtimetocollisionestimationautonomous} propose a two-stage method that first employs a robust linear solver based on a time-variant affine model, followed by further refinement of the resulting model through spatio-temporal registration.
In contrast to the aforementioned methods, AEB-Tracker~\cite{wang2024asynchronous} uses an extended Kalman filter to track the state of event blobs (\textit{e.g.}, the tail lights of vehicles), thereby obtaining the scale variation of the target object.

\textbf{2) \textit{Autonomous Driving Datasets Focused on Collision Risk
Prediction and TTC Estimation Tasks}}:
Currently, numerous datasets exist for collision risk prediction, such as DR(eye)VE~\cite{palazzi2018predicting} and DADA-2000~\cite{fang2019dada}, both of which offer data from a variety of scenarios. 
For example, DR(eye)VE provides over 500k frames of normal driving scenes with driver eye-tracking data.
Moreover, \cite{kim2019crash} offers 2D bounding boxes and collision labels for emergency braking and normal driving scenes. 
However, these datasets lack ground truth of depth and trajectory, limiting their usage for TTC estimation. 
The TSTTC~\cite{shi2023tsttc} dataset, which focuses on TTC estimation, contains over 200k frames with ground-truth TTC and bounding boxes.
Nevertheless, these datasets only include scenarios under normal driving conditions and lack high-speed emergency braking situations, which are crucial for testing TTC algorithms under extreme conditions.

\textbf{3) \textit{Event Camera-Centric Datasets}}:
There are a number of event camera-centric datasets that include driving scenarios.
Among them, MVSEC~\cite{zhu2018multivehicle} is the first stereo event camera dataset, where the stereo rig is handheld or mounted on a mobile platform, such as a drone, an automobile and a motorcycle.
Afterwards, Gehrig \etal \cite{Gehrig21ral} present a large-scale stereo event dataset with higher spatial resolution, better sensor synchronization, and a wide baseline, leading to better depth estimation accuracy. 
To handle extreme lighting conditions, ViViD++\cite{lee2022vivid++} integrates thermal sensors alongside event cameras, but hardware triggered synchronization is not witnessed.
Limited by the immaturity of event cameras at that time, high definition (HD) data are not available in these datasets until the appearance of M3ED\cite{chaney2023m3ed}, which utilizes a Prophesee EVK4 event camera with a spatial resolution of 1280$\times$720 pixels.
In general, these datasets lack specific scenarios and data for the development of TTC tasks.

Different from above-mentioned datasets, the proposed one (called EvTTC) focuses on providing pre-collision data across various speeds and scenarios.
The data are collected using a hardware-synchronized sensor suite, which includes two pairs of high-resolution event and RGB cameras with different focal length for covering a larger and wider field of view.
Besides, the sensor suite also consists of a LiDAR, an IMU, and a GNSS/INS for providing high-frequency ground-truth information.
The pre-collision data involve vehicles and pedestrians, offering comprehensive ground truth of depth, vehicle's pose, TTC, and 2D bounding boxes of front obstacles.
As a result, it will fill the gap in high-quality data for TTC tasks under challenging driving conditions.

%% file: chapters/03_Hardware_Setup.tex
\section{Hardware Configuration}
\label{sec: hardware}
The setup of the hardware is discussed in this section.
Specifically, we disclose the sensors and devices used for recording raw data and ground truth (Sec.~\ref{subsec:sensors}), followed by the details of time synchronization across different devices (Sec.~\ref{subsec:time synchronization}) and multi-sensor calibration (Sec.~\ref{subsec:calibration}).

%++++++++++++++++++++++++++++++++++++++++++++++++++++++++++++++++
\subsection{Sensors}
\label{subsec:sensors}
\input{floats/tab_sensor_setup}
\input{floats/fig_sensors_synchronation}
\input{floats/fig_calibr_sensors}
\input{floats/fig_top_view}
\input{floats/fig_real_scene}
\input{floats/tab_fig_seq_params}
%++++++++++++++++++++++++++++++++++++++++++++++++++++++++++++++++

The raw data used for estimating TTC contain RGB images and corresponding event streams.
To cover a relatively complete sensing range, we employ two groups of RGB-Event camera pair equipped with lens of different focal length.
The camera pair using 8-mm lens consists of an RGB camera and an event camera, and it covers the close sensing range (see \circled{A} in Fig.~\ref{fig:eye_catcher}).
As a complementary, the camera pair using 16-mm lens is used to cover the distant sensing range (see \circled{B} in Fig.~\ref{fig:eye_catcher}).
In each pair, the event camera and the RGB camera are rigidly attached with a narrow baseline of 4 cm, and thus, the extrinsic parameters (w.r.t the coordinate system of the sensor suite) can be approximately shared, simplifying data fusion across the two sensor modalities.
Detailed parameters of each device can be found in Tab.~\ref{tab: sensor_setup}.

The ground-truth data consists of depth information observed from the host vehicle and the involved two vehicles' motion status (\eg, position, speed, and orientation, etc).
To provide these information, we employ a solid-state LiDAR and two high-performance GNSS/INS units (see Tab.~\ref{tab: sensor_setup}). 
% \textcolor{blue}{\sout{A Livox HAP LiDAR is rigidly attached to the two camera pairs and used for collecting ground-truth depth. 
% As for the GNSS/INS devices, a UG005 X1 device is used on the host vehicle, and a CGI-610 device is used on the lead vehicle, both for providing high-frequency motion status of the platform.}
% }
Note that international organizations of vehicle safety assessment, such as Euro New Car Assessment Program (NCAP), require the motion status of involved vehicles to be collected at a frequency of no lower than 100 Hz. 
To this end, both GNSS/INS units are configured to provide motion status at 100 Hz.

\subsection{Time Synchronization}
\label{subsec:time synchronization}
To this end, we utilize Precision Timing Protocol (PTP)~\cite{ieee1588-2019} and the generalized Precision Time Protocol (gPTP) \cite{ieee8021as2020}, which can provide sub-microsecond synchronization accuracy in the Ethernet.

The time synchronization among all the devices is illustrated in Fig.~\ref{fig: sensors_synchronation}.
First, we use a PTP server to synchronize GNSS's clock with the system clock of the industrial personal computer (IPC).
The IPC is then designated as the master clock, using the PTP protocol to synchronize with the RGB cameras. 
Since the Livox HAP only supports the gPTP protocol, we use gPTP to synchronize between the IPC and the LiDAR.
To synchronize the event cameras with the system time, we utilize a micro-controller to generate four synchronization pulses at 20 Hz. 
These pulses are used to simultaneously trigger the two RGB cameras and the two event cameras.
% \textcolor{blue}{\sout{The RGB and event cameras are configured in external trigger mode.
% When the RGB camera receives a synchronization pulse, it begins to capture an image and record its timestamp. 
% Meanwhile, the event camera would also be triggered by the synchronization pulse, immediately generating a signal with a timestamp. 
% The temporal offset between the RGB camera and the event camera can be easily calculated as the difference between the two timestamps.}}
% \jhang{Check ESVO2's rebuttal and decide whether to show the modification in our rebuttal.}
In addition, the integrated navigation systems UG005 X1 and CGI-610 are synchronized via their built-in GNSS clocks.

\subsection{Calibration}
We discuss all involved calibration tasks in this part, including the cameras' intrinsic and extrinsic calibration, extrinsic calibration between cameras and the IMU, and that between cameras and the LiDAR.
The precise CAD model of our sensor suite is illustrated in Fig.~\ref{fig: calibr_sensors}.
All calibration results are made available in both YAML and HDF5 formats.

\label{subsec:calibration}
\subsubsection{Intrinsic and Extrinsic Calibration of Cameras}
We use the Kalibr toolbox~\cite{furgale2013unified} to estimate the intrinsic and extrinsic parameters for the two camera pairs separately. 
Since the Kalibr toolbox requires image-format data for camera calibration, we utilize the Simple Image Reconstruction library~\footnote{ \scriptsize \url{https://github.com/berndpfrommer/simple_image_recon}} to recover frames from raw event data.
For accurate extrinsic parameters estimation, we first synchronize the event and RGB cameras using the synchronization scheme (Sec.~\ref{subsec:time synchronization}), and then perform image reconstruction at the midpoint of each image's exposure time.
\subsubsection{Extrinsic Calibration of Cameras and IMU}
For the extrinsic calibration between the cameras and the IMU, we effectively stimulate the sensor suite along the six degrees of freedom of the IMU in front of an \textit{AprilTag} grid pattern and record the corresponding sequences.
The Kalibr toolbox is then used to perform the extrinsic calibration.

\subsubsection{Extrinsic Calibration of Cameras and LiDAR}
The extrinsic calibration between the RGB cameras and the LiDAR is performed as follows.
We first keep the sensor suite static in a structured outdoor scene, and then record multiple point cloud frames along with the corresponding RGB images.
The initial guess for the calibration is provided using the CAD statistics of the sensor suite, followed by fine-tuning with the Sensors Calibration toolbox~\cite{opencalib} that refines the extrinsic parameters.

%% file: floats/tab_sensor_setup.tex
\begin{table}[t]
\setlength{\tabcolsep}{3pt}
\centering
\resizebox{0.45\textwidth}{!}{%
\begin{tabular}{l l l}
    \toprule
      \makecell*[c]{Devices} &  Models & Parameters \\
     \midrule
     \makecell*[c]{8-mm Lens \\ Camera Pair} & \makecell[l]{Prophesee EVK4 \vspace{0.35cm}\\ \\FLIR Blackfly S} & \makecell[l]{1280$\times$720 1/2.5'' \\ FoV: 41°$\times$24 \\ \midrule 1920$\times$1200 1/2.3" \\ FoV: 44°$\times$29° \\ Rate: 20Hz}\\
     \midrule
     \makecell*[c]{16-mm Lens \\ Camera Pair} & 
     \makecell[l]{Prophesee EVK4 \vspace{0.35cm}\\ \\FLIR Blackfly S} & 
     \makecell[l]{1280$\times$720 1/2.5'' \\ FoV: 22°$\times$12° \\ 
     \midrule1920$\times$1200 1/2.3" \\ FoV: 23°$\times$14° \\ Rate: 20Hz}\\
     \midrule
     \makecell*[c]{LiDAR} & Livox HAP (TX) & \makecell[l]{150m @ 10\% NIST \\ ±3cm range accuracy \\ FoV: 120°$\times$25° \\ IMU: BMI088}\\
     \midrule
     \makecell*[c]{GNSS/INS} & \makecell[l]{CGI-610 \\ \\ UG005 X1 } & \makecell[l]{Dual-Antenna RTK\\ Rate: 100Hz \\ ±3cm pos. accuracy \\ ±2cm/s vel. accuracy} \\
     \bottomrule 
\end{tabular}
}
\caption{Setup and parameters of the sensor suite.}
% \vspace{-1.5 cm}
\label{tab: sensor_setup}
\end{table}

%% file: floats/fig_sensors_synchronation.tex
\begin{figure}[t]
    \centering
    \includegraphics[width=0.9\linewidth]{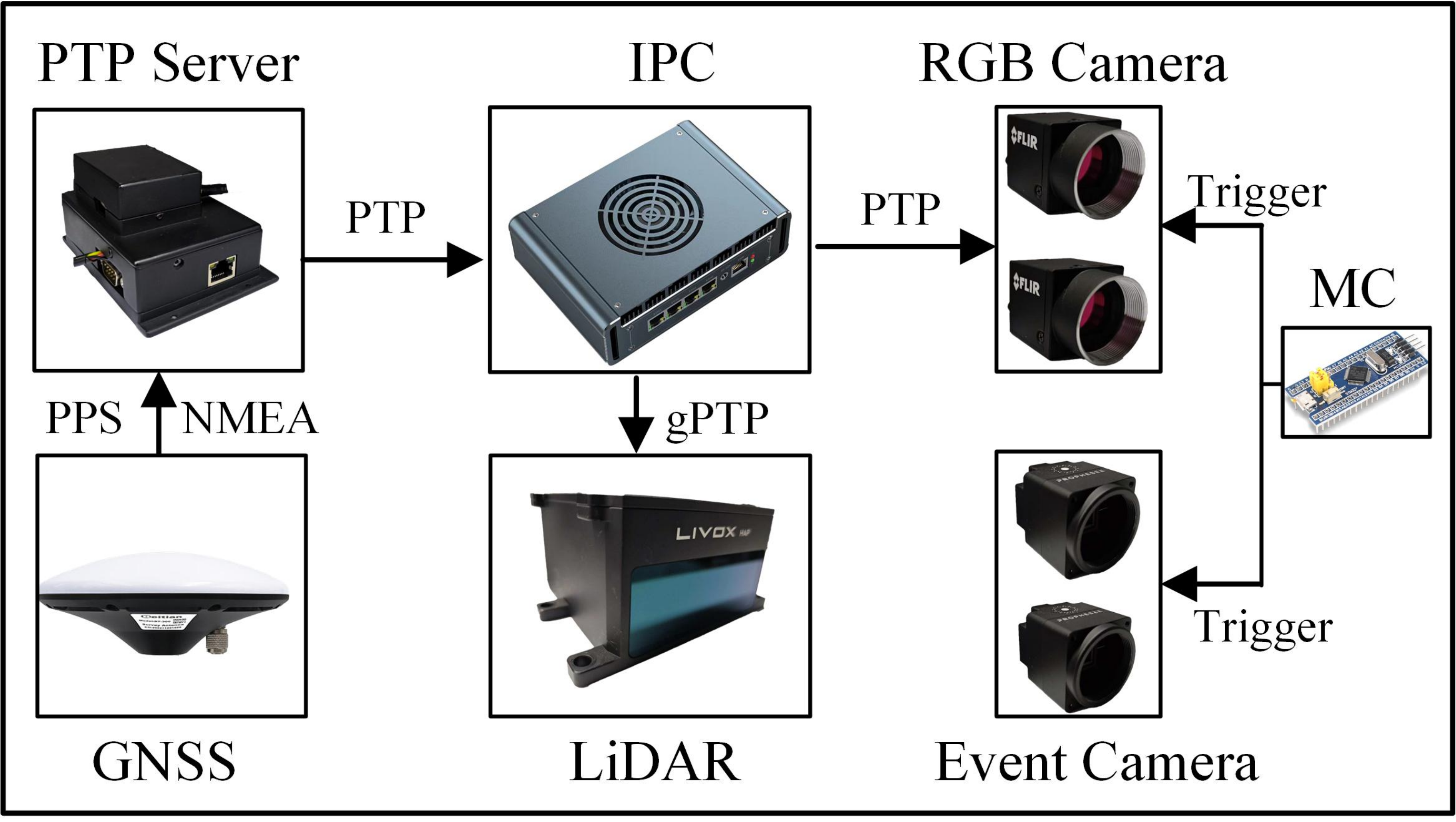}
    \caption{Illustration of our synchronization scheme.}
    \label{fig: sensors_synchronation}
    \vspace{-2.5cm}
\end{figure}

%% file: floats/fig_calibr_sensors.tex
\begin{figure}[t]
    \centering
    \includegraphics[width=0.8\linewidth]{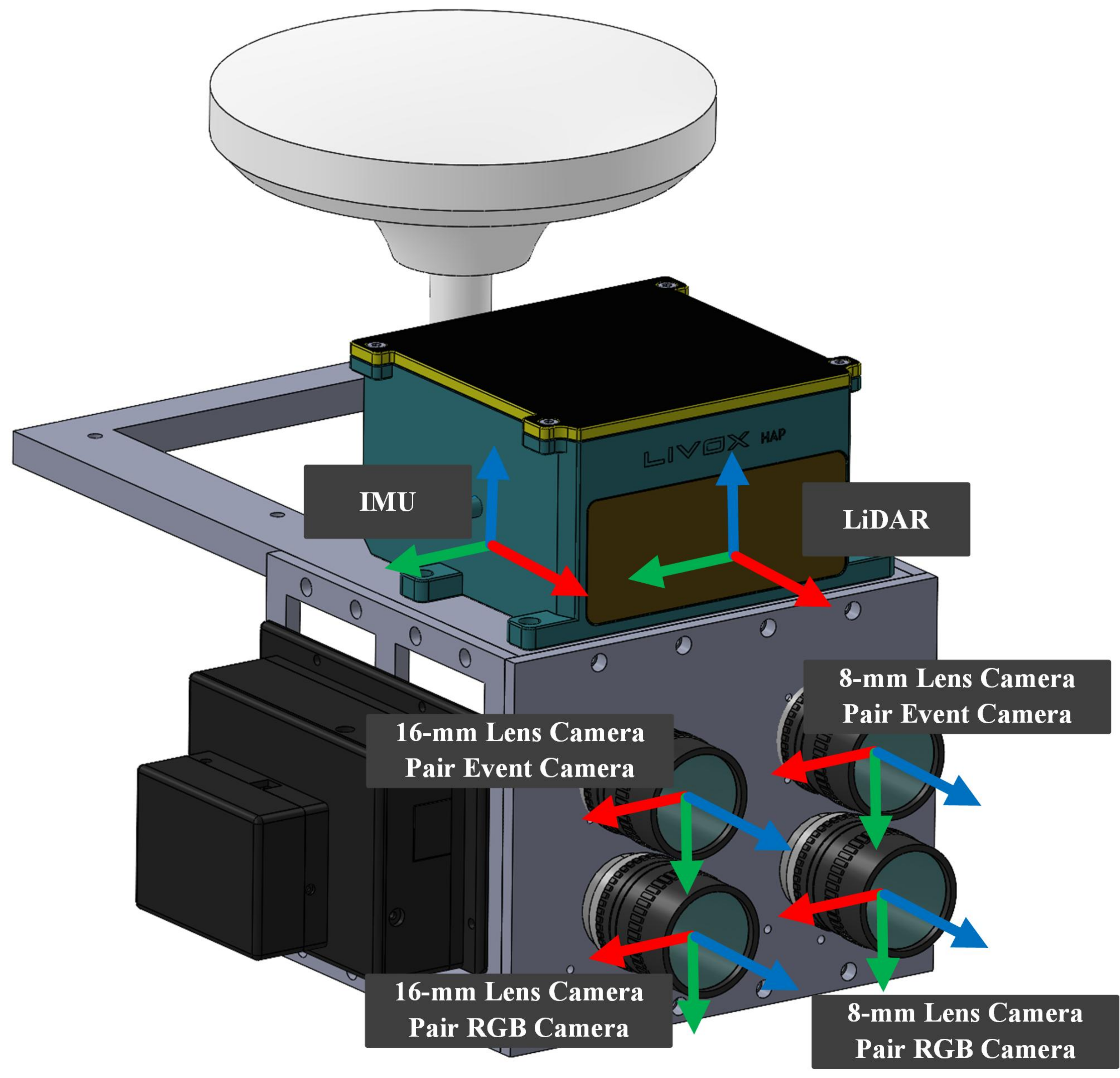}
    \caption{
    \textit{Illustration of the CAD model of the sensor suite.} The axes of all sensors are labeled and color-coded as follows: red for X, green for Y, and blue for Z.}
    \label{fig: calibr_sensors}
    % \vspace{-2 cm}
\end{figure}

%% file: floats/fig_top_view.tex
\begin{figure*}[ht]
    \centering
    \includegraphics[width=0.95\linewidth]{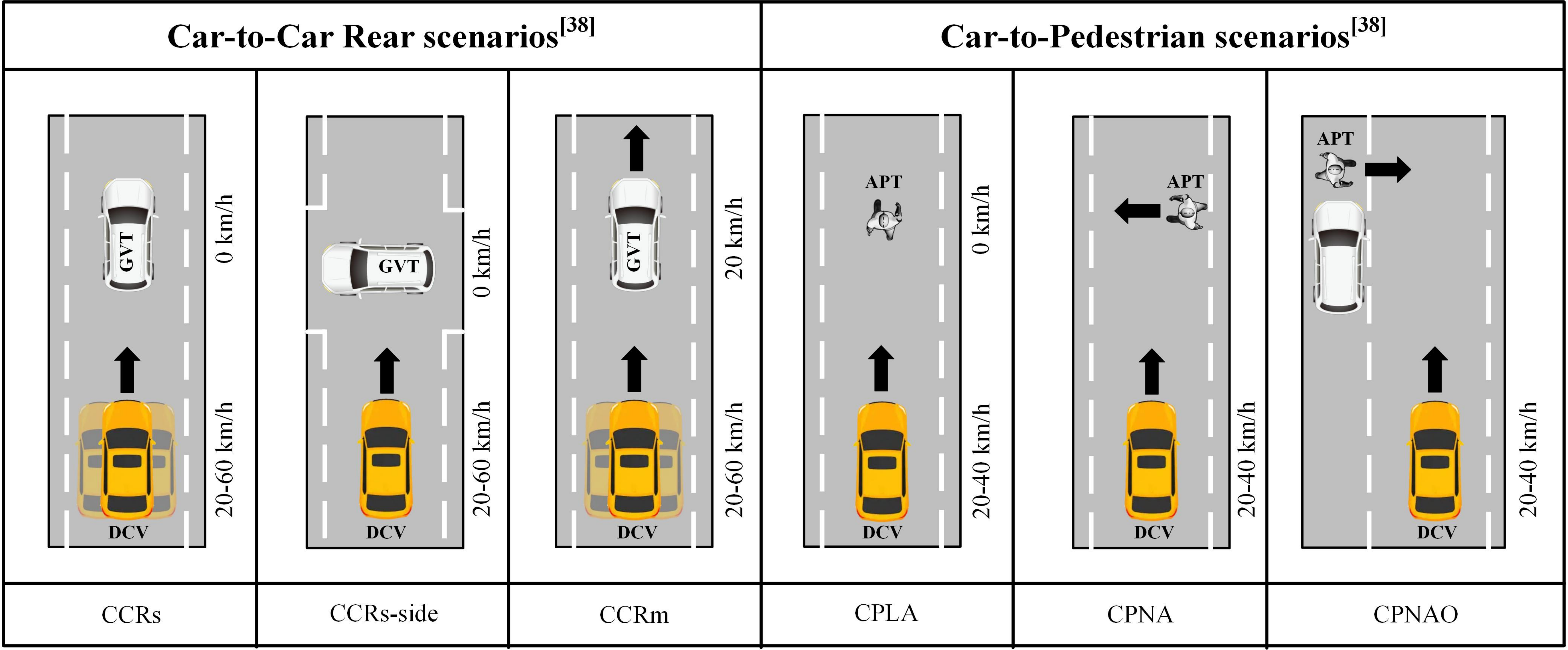}
    \caption{\textit{The top-view schematic of the dataset scenarios.} The arrow represents the direction of movement. The lateral shadowing of the DCV in the CCRs and CCRm scenarios indicates that data are collected across different lane positions. The objects in the scene include the Global Vehicle Target (GVT), Adult Pedestrian Target (APT), and Data Collection Vehicle (DCV).}
    \label{fig:top-view}
\end{figure*}

%% file: floats/fig_real_scene.tex
\begin{figure*}[ht]
\centering
\resizebox{0.95\textwidth}{!}{
\begin{minipage}{\textwidth}

\subfloat[CCRs-1]{\includegraphics[width=0.24\textwidth]{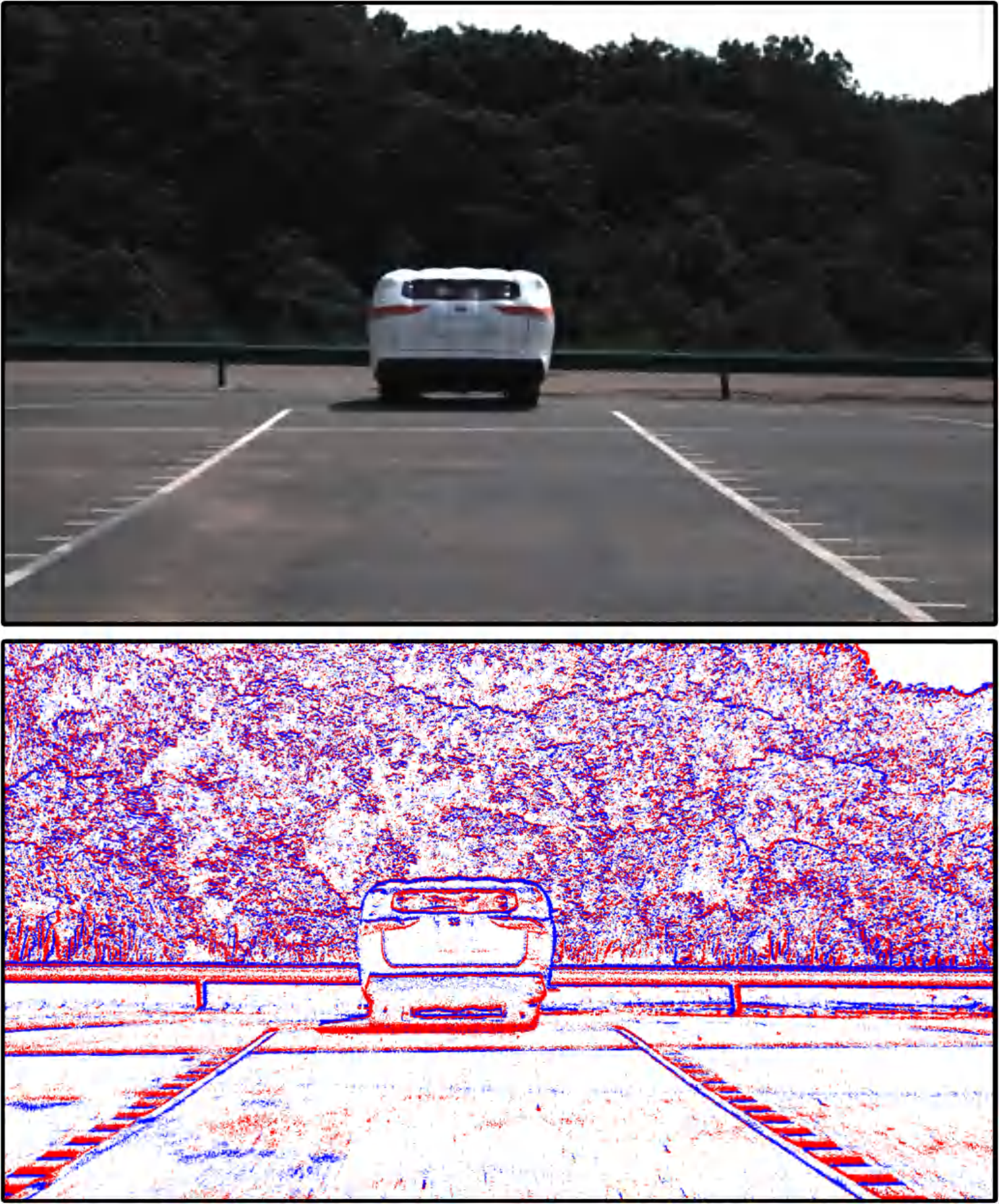} \label{fig:ccrs-1}}
\hfill 	
\subfloat[CCRs-2]{\includegraphics[width=0.24\textwidth]{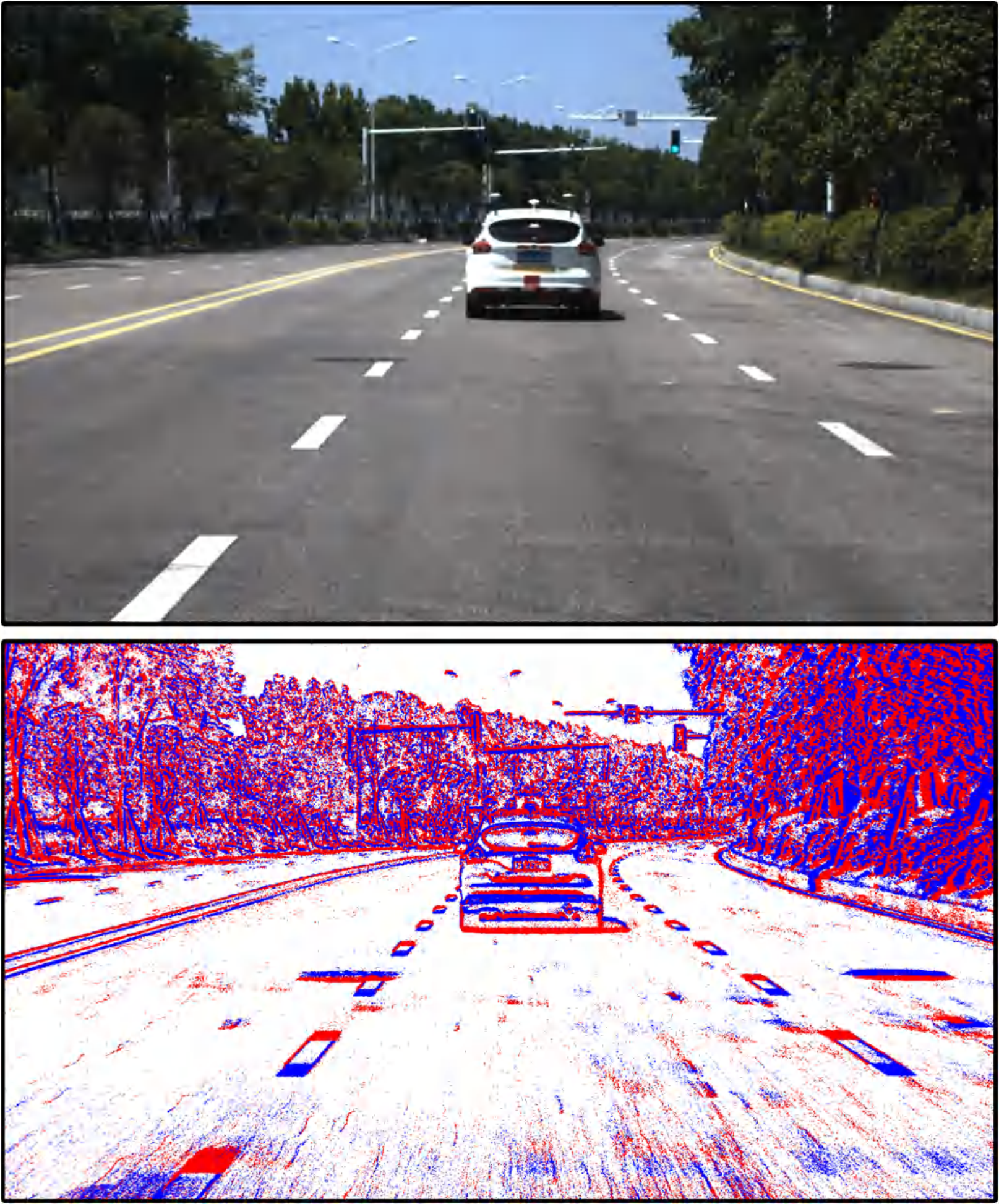} \label{fig:ccrs-2}}
\hfill 
\subfloat[CCRs-3]{\includegraphics[width=0.24\textwidth]{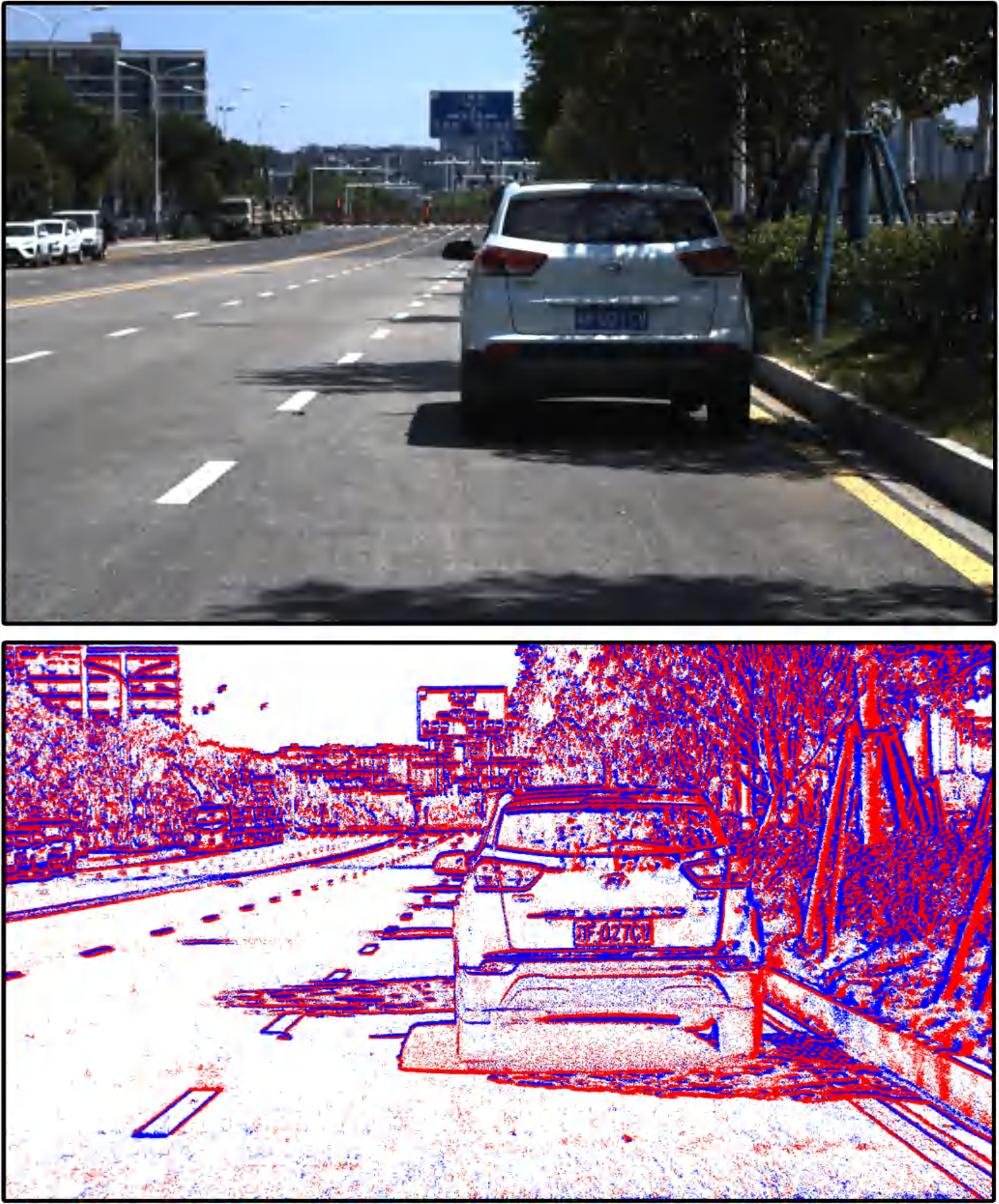} \label{fig:ccrs-3}}
\hfill 
\subfloat[CCRs-side]{\includegraphics[width=0.24\textwidth]{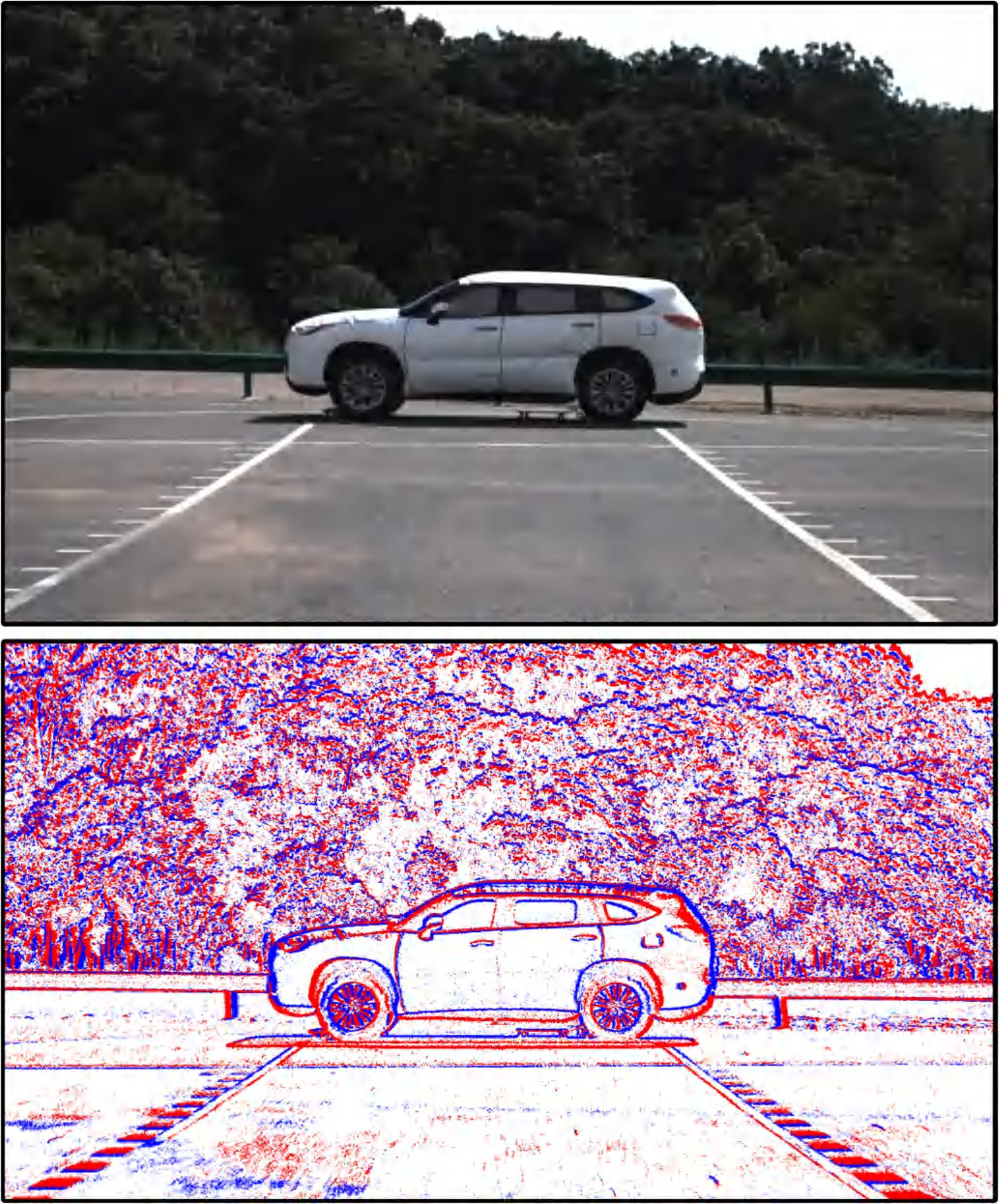} \label{fig:ccrs-side}}
\hfill 
\subfloat[CCRm]{\includegraphics[width=0.24\textwidth]{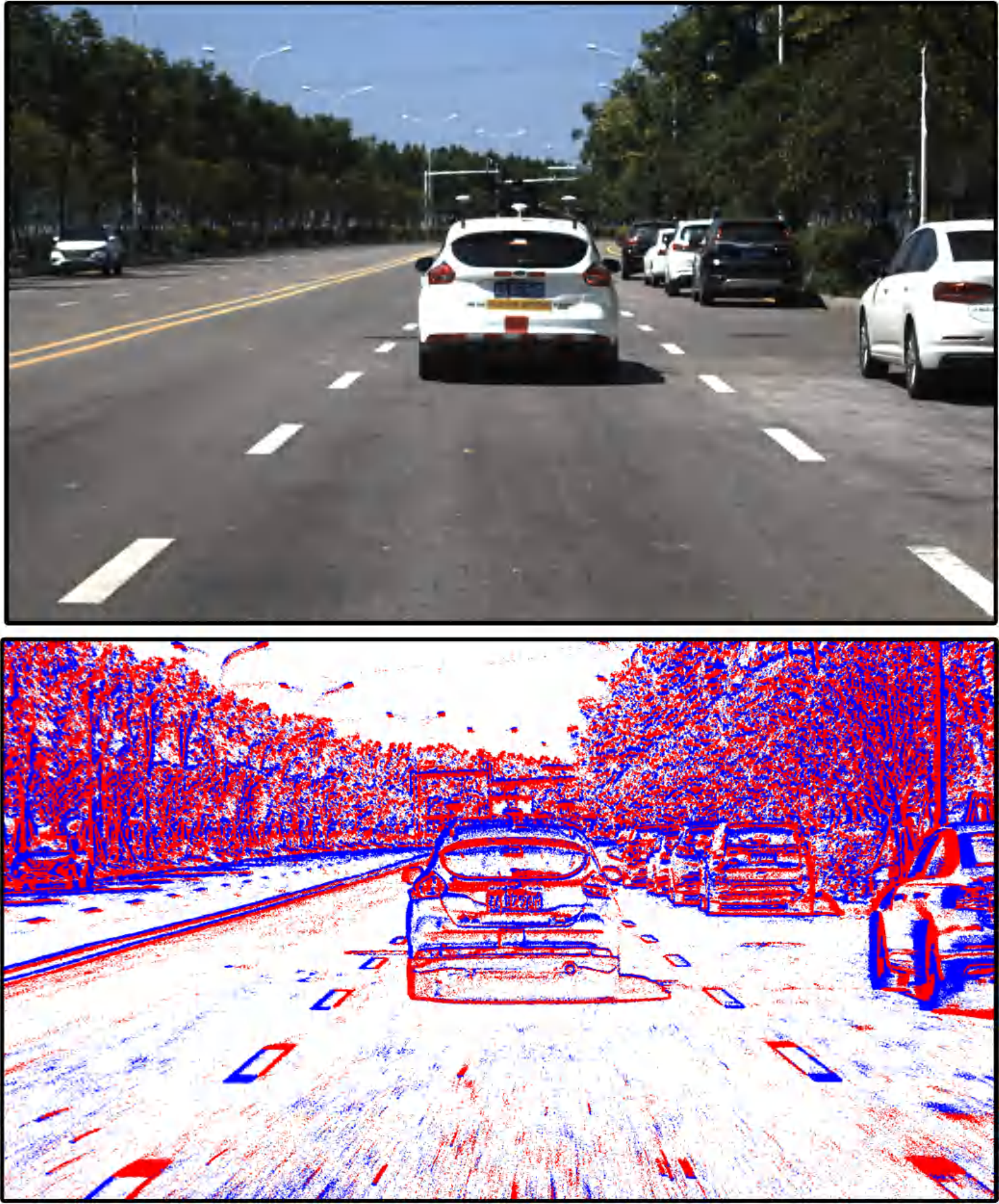} \label{fig:ccrs-m}}
\hfill 	
\subfloat[CPLA]{\includegraphics[width=0.24\textwidth]{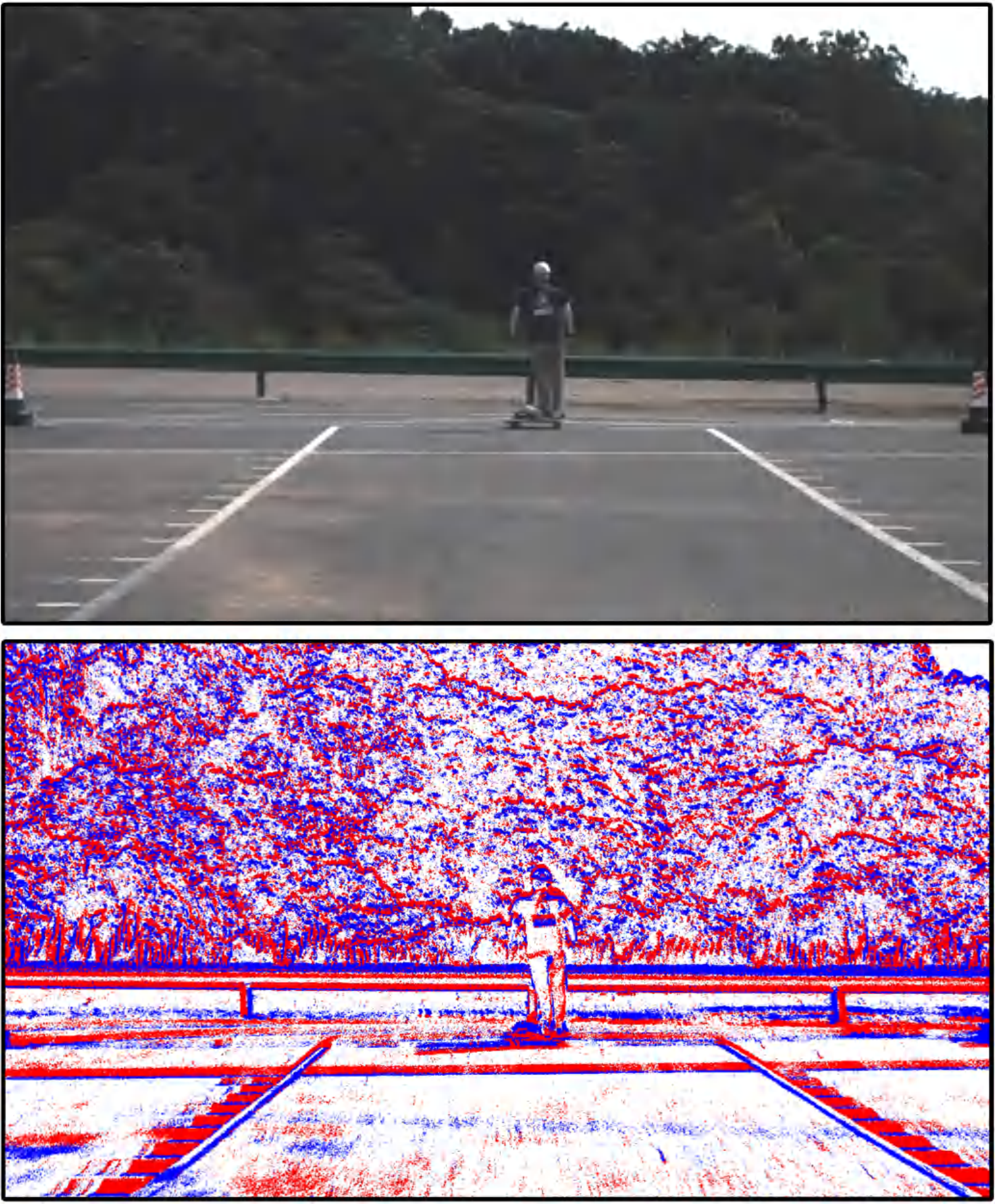} \label{fig:cpla}}
\hfill 
\subfloat[CPNA]{\includegraphics[width=0.24\textwidth]{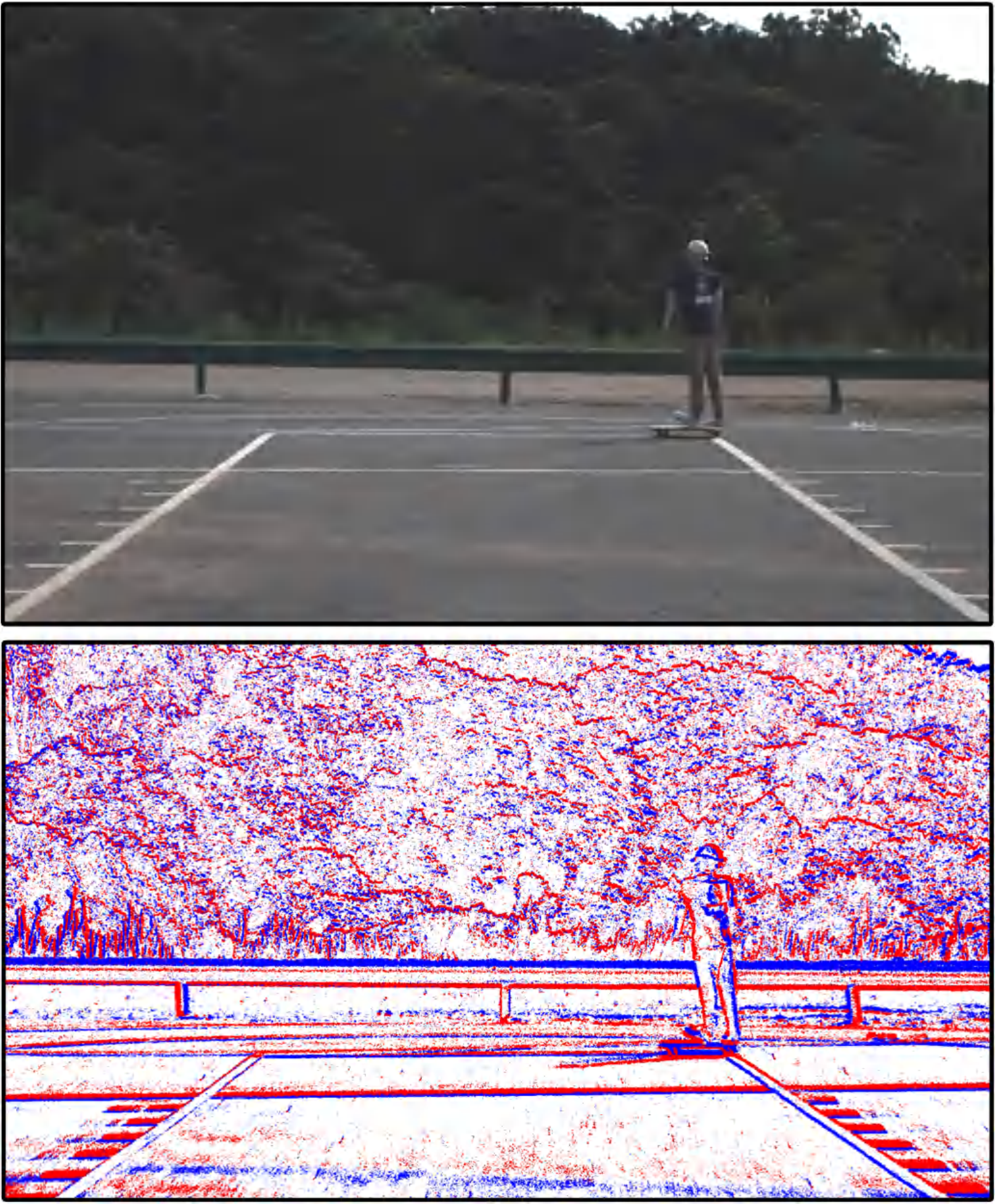} \label{fig:cpna}}
\hfill 
\subfloat[CPNAO]{\includegraphics[width=0.24\textwidth]{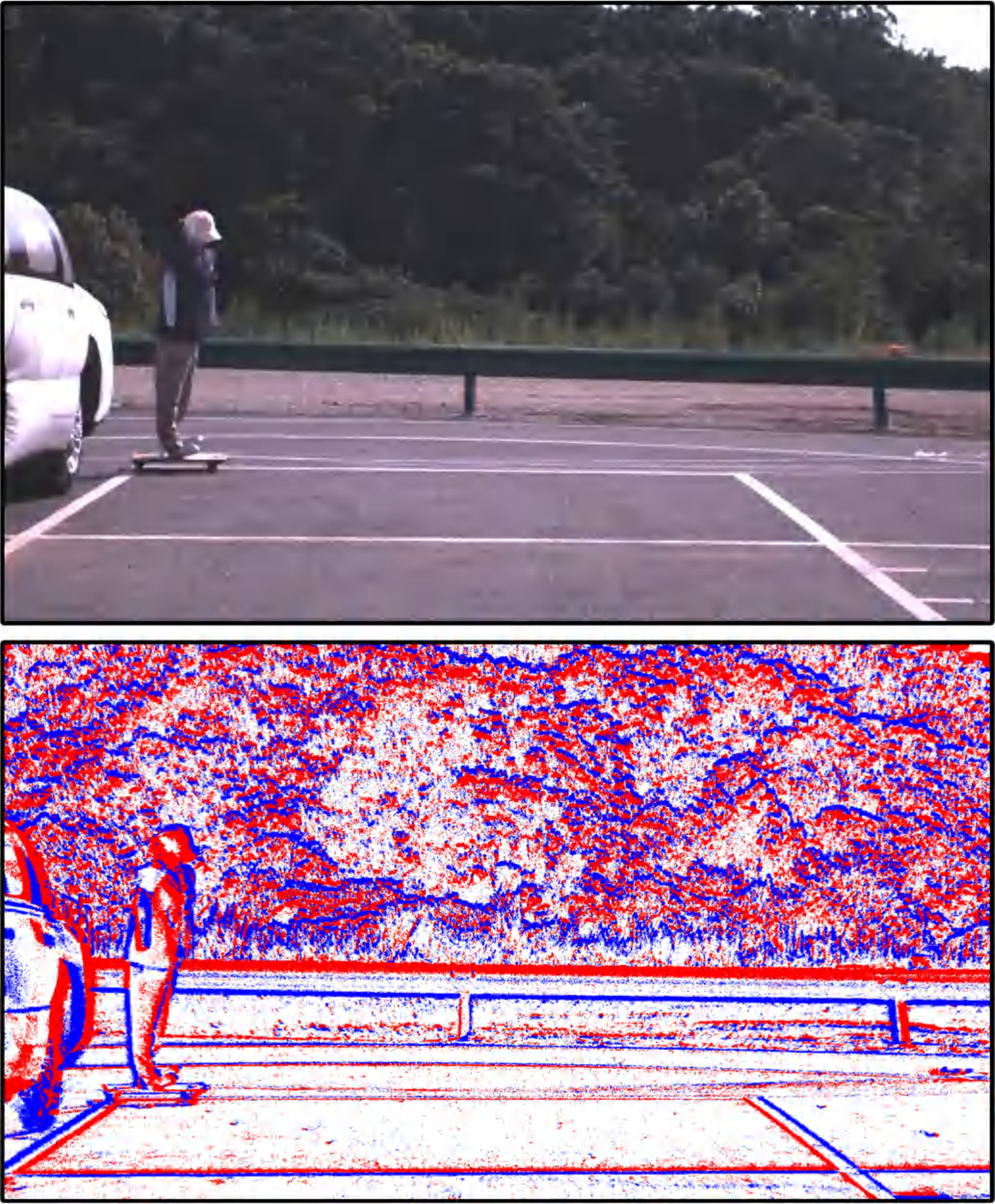} \label{fig:cpnao}}

\end{minipage}
}
\hfill 
    \caption{\textit{Illustration of real-world road scenes in our dataset.}
    The top of the first and second rows respectively shows the RGB images for each scenario, while the bottom of the first and second rows respectively presents the accumulated event data for each scenario.
    }
    \label{fig:real-scene}
    \vspace{-0.5cm}
\end{figure*}

%% file: floats/tab_fig_seq_params.tex
\begin{figure*}[t]
    \centering
    \begin{minipage}{0.38\textwidth}
        \centering
        \includegraphics[width=\textwidth]{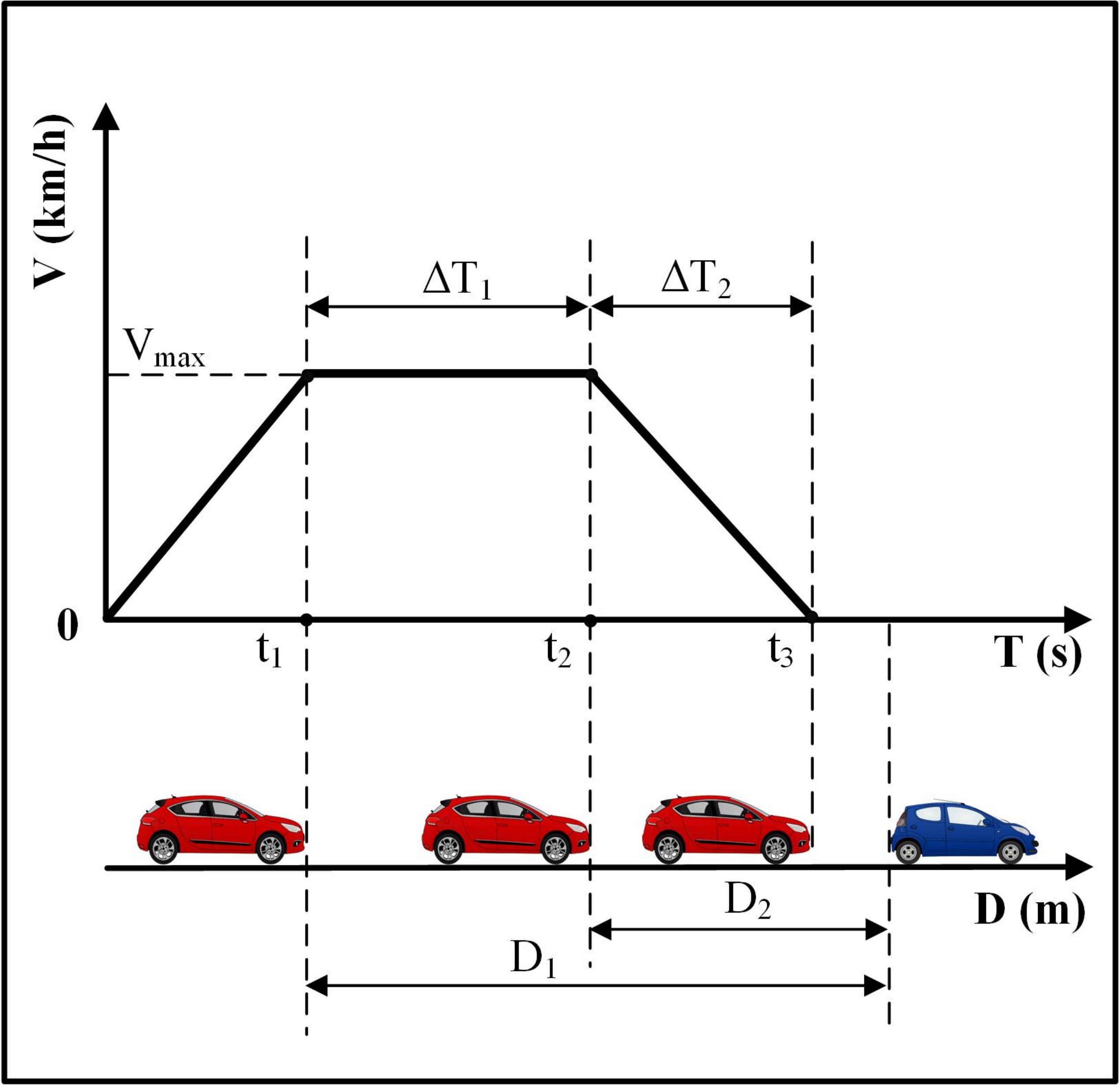}
        \caption{\textit{Parameter definitions in TTC Scenarios.} $\Delta \text{T}_1\text{(s)}$: duration at constant speed, $\Delta \text{T}_2\text{(s)}$: braking duration, $\text{D}_1\text{(m)}$: distance to the collision target when the DCV reaches maximum speed, $\text{D}_2\text{(m)}$: distance to the collision target at braking onset.}
        \label{fig:ttc_process}
    \end{minipage}
    \hspace{0.04\textwidth}
    \begin{minipage}{0.5\textwidth}
    \centering
    \begin{adjustbox}{max width=\linewidth}
        \setlength{\tabcolsep}{3pt}
        \begin{tabular}{lcccccccc}
            \toprule[1.5pt]
            Seq. Name & $V_{rel}$ Tier   & $\|V_{rel}\|_{max}$(km/h)   & $\text{D}_1\text{(m)}$ & $\text{D}_2\text{(m)}$  & $\Delta \text{T}_1\text{(s)}$  & $\Delta \text{T}_2\text{(s)}$  & Total  \\  
            \midrule
            \multirow{3}{*}{CCRs-1}    & low               & 22.7                      & 50        & 7         & 7.4             & 1.3              & 3       \\
                                       & medium            & 33.5                      & 32        & 9         & 2.5              & 1.5              & 3      \\ 
                                       & high              & 70.4                      & 109       & 31        & 4.2             & 3.5              & 3       \\
            \midrule
            \multirow{3}{*}{CCRs-2}    & low               & 14.2                      & 35        & 12        & 4.0              & 3.0              & 1       \\
                                       & medium            & 40.3                      & 58        & 24        & 3.0              & 4.0              & 1       \\
                                       & high              & 57.6                      & 115       & 38        & 4.8              & 5.5              & 1       \\
            \midrule
            \multirow{2}{*}{CCRs-3}    & low               & 24.1                       & 31        & 14        & 2.6              & 3.5              & 1       \\
                                       & medium            & 41.4                      & 42        & 25        & 1.5              & 3.8              & 1       \\
            \midrule
            \multirow{3}{*}{CCRs-side} & low               & 21.2                      & 46        & 6         & 7.0              & 1.1              & 1       \\
                                       & medium            & 35.3                      & 42        & 12        & 3.0              & 2.0              & 1       \\
                                       & high              & 61.9                      & 76        & 40        & 2.1              & 4.5              & 1       \\
            \midrule
            \multirow{2}{*}{CCRm}      & low               & 25.2                      & 57        & 20        & 5.5              & 3.5              & 3       \\
                                       & medium            & 37.4                      & 34        & 19        & 1.5              & 3.0              & 3       \\
            % \hhline{========}
            \midrule
            \multirow{3}{*}{CPLA}      & low               & 24.5                      & 88        & 6         & 13.1             & 1.3              & 1       \\
                                       & medium            & 41.4                      & 89        & 10        & 7.1              & 1.7              & 1       \\
                                       & high              & 61.6                      & 106       & 26        & 4.9              & 3.5              & 1       \\
            \midrule
            \multirow{3}{*}{CPNA}      & low               & 23.8                      & 92        & 11        & 13.0             & 1.9              & 1       \\
                                       & medium            & 38.9                      & 108       & 12        & 9.1              & 1.6              & 1       \\
                                       & high              & 61.7                      & 108       & 23        & 5.2              & 2.5              & 1       \\
            \midrule
            \multirow{3}{*}{CPNAO}     & low               & 23.7                      & 94        & 8         & 14.2             & 1.9              & 1       \\
                                       & medium            & 39.7                      & 104       & 18        & 8.2              & 2.5              & 1       \\
                                       & high              & 55.0                        & 87        & 42        & 2.9              & 4.7              & 1       \\
            \bottomrule[1.5pt]
        \end{tabular}
    \end{adjustbox}
    
        \captionof{table}{The motion parameters in each TTC scenario.}
        \label{tab:sequences_list}
        % \vspace{-0.5cm}
    \end{minipage}
\end{figure*}

%% file: chapters/04_Sequences_Overview.tex
\section{Sequences Overview}
\label{sec: sequences_overview}
This section presents a comprehensive overview of the dataset.
The scenario configurations and vehicle motion conditions are developed based on the Car-to-Car and Car-to-Pedestrian emergency collision scenarios defined in the Euro NCAP AEB test protocol~\cite{euroNCAP2023AEB}. 
For further details, please refer to Sec.~\ref{Scenarios} and the illustrations in Fig.\ref{fig:top-view} and Fig.~\ref{fig:real-scene}.
The ground-truth TTC, vehicle poses, and depth information are 
generated by the data from LiDAR and GNSS/INS units (Sec.~\ref{Ground-truth Generation}).

\subsection{Scenarios}
\label{Scenarios}
\subsubsection{Car-to-Car Rear scenarios}
We collect data from two types of scenarios: Car-to-Car Rear stationary (CCRs) and Car-to-Car Rear moving (CCRm). 
We use two types of real vehicles and a 1:1 scale inflatable vehicle as collision targets, with the latter intended to simulate more realistic collision scenarios while ensuring safety.
In the CCRs, when the GVT is a real vehicle, it is placed only longitudinally, as shown in Fig.~\ref{fig:real-scene}\subref{fig:ccrs-2} and Fig.~\ref{fig:real-scene}\subref{fig:ccrs-3}. 
When the GVT is an inflatable vehicle, it is placed both longitudinally and laterally, as illustrated in  Fig.~\ref{fig:real-scene}\subref{fig:ccrs-1} and Fig.~\ref{fig:real-scene}\subref{fig:ccrs-side}. 
In the CCRm, the GVT moves at a constant speed of 20 km/h (see Fig.~\ref{fig:real-scene}\subref{fig:ccrs-m}).
Additionally, in the CCRs-1 (Fig.~\ref{fig:real-scene}\subref{fig:ccrs-1}) and CCRm, the DCV and the GVT occupy different lane positions, including fully aligned, partially offset, and completely offset cases.
In all scenarios, the DCV approaches the GVT at speeds ranging from approximately 20 to 60 km/h, as detailed in Fig.~\ref{fig:ttc_process} and Tab.~\ref{tab:sequences_list}.

\subsubsection{Car-to-Pedestrian scenarios}
We collect data from three types of scenarios: Car-to-Pedestrian Longitudinal Adult (CPLA), Car-to-Pedestrian Nearside Adult (CPNA), and Car-to-Pedestrian Nearside Adult Obstructed (CPNAO). 
We select an inflatable dummy as the APT.
In the CPLA, the APT is placed in the middle of the road, facing away from the DCV, as shown in Fig.~\ref{fig:real-scene}\subref{fig:cpla}. 
In the CPNA, the APT crosses the road at a constant speed, as illustrated in Fig.~\ref{fig:real-scene}\subref{fig:cpna}. 
In the CPNAO, the APT suddenly crosses the road from outside the DCV's field of view (see Fig.~\ref{fig:real-scene}\subref{fig:cpnao}). 
In all scenarios, the DCV approaches the APT at speeds ranging from approximately 20 to 60 km/h, as shown in Fig.~\ref{fig:ttc_process} and Tab.~\ref{tab:sequences_list}.

\subsection{Ground-truth Generation}
\label{Ground-truth Generation}
\subsubsection{TTC}
In the camera coordinate system, the TTC can be calculated as follows:
\begin{equation}
    \text{TTC} = \frac{Z}{V_{rel}}
    \label{eq:ttc_formula}
\end{equation}
where $Z$ represents the depth of the collision target in the camera coordinate system, and $V_{rel}$ denotes the relative velocity between the DCV and the collision target along the camera's optical axis.

The ground-truth data for depth and vehicle motion status in all sequences are obtained through LiDAR and GNSS/INS units, respectively (see Table~\ref{tab: sensor_setup}). 
Specifically, the motion status data for the DCV are collected using the UG005 X1 unit, while the CGI-610 unit is used to collect motion status data for the GVT only in the CCRm scenario, as illustrated in Fig.~\ref{fig:real-scene}\subref{fig:ccrs-m}.
In other scenarios, since the collision target does not move along the camera's optical axis, the position of the collision target can be determined by measuring the distance between the DCV and the target using LiDAR.

With the calibrated extrinsic parameters, all sequence data are transformed into the RGB camera coordinate frame of the 8-mm lens camera pair.
The ground-truth TTC is then calculated using the relative depth $Z$ and relative velocity $V_{rel}$ by Eq.~\ref{eq:ttc_formula}.

\subsubsection{Pose and Depth}
We use the LiDAR odometry FAST-LIO2~\cite{xu2022fast} to obtain ground-truth poses and velocity-compensated point clouds.
For each pose, we extract the relative poses of several preceding and succeeding point cloud frames to generate local point clouds.
Next, the Hidden Point Removal (HPR) operator~\cite{katz2007direct} is applied to eliminate occlusion effects in the local point clouds. 
Finally, the local point clouds are projected into the event camera coordinate of the 8-mm and 16-mm lens camera pairs to obtain ground-truth depth.

%% file: chapters/Collision_Simulation_SLIDER.tex
\section{SMALL-SCALE TESTBED}
\input{floats/fig_slider_schematic}
\input{floats/tab_hybird_optical_system}
\label{sec: collision simulation slider}
\input{floats/tab_benchmark}

Considering the gap between synthetic data and real data, the latter is indispensable in the verification of algorithms.
However, online evaluation in real-world scenes can be resource intensive.
Thus, a design of small-scale TTC testbed, for simulating the emergency scenarios of vehicle collision, is presented.

The testbed consists of a hybrid optical system~\cite{hidalgo2022event}, a linear rail and a 1:24 scale model vehicle. 
Fig.~\ref{fig:slider-schematic} illustrates the platform setup, and Tab.~\ref{tab:hybird_optical_system} specifies the device parameters.
Specifically, the hybrid optical system comprises an DVXplorer event camera, an RGB camera, and a beam splitter. 
The beam splitter divides the incoming light into two paths, ensuring that both cameras share a unified field of view, facilitating pixel-level correspondence between the event and RGB cameras. 
To achieve precise time synchronization, a micro-controller is used to generate two synchronized 25 Hz pulse signals, which are employed to simultaneously trigger both cameras. 
The linear rail has an effective travel distance of 2 meters, with a control system that includes a servomotor, a driver, and an STM32 development board.
Besides, a 1:24 scale model vehicle is positioned in front of the camera, being used as the potential collision target. 
The hybrid optical system is rigidly mounted on the slider, which can be propelled towards the target under accurate control of speed.
The ground-truth TTC is calculated based on the velocity and relative displacement measured by the motor encoder.

%% file: floats/fig_slider_schematic.tex
\begin{figure*}[t]
    \centering
    \includegraphics[width=1\linewidth]{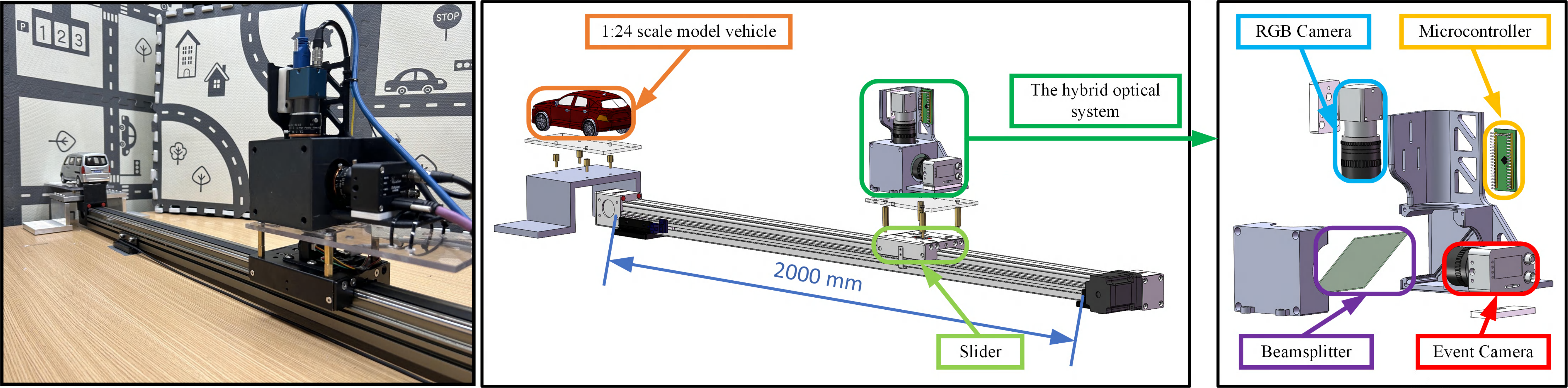}
    \caption{From left to right: the small-scale TTC testbed, the assembly diagram, and the layout of the hybrid optical system.}
    \label{fig:slider-schematic}
\end{figure*}

%% file: floats/tab_hybird_optical_system.tex
\begin{table}[tb]
    \setlength{\tabcolsep}{3pt}
    \centering
    \resizebox{0.45\textwidth}{!}{%
    \begin{tabular}{l l l}
        \toprule
        \makecell*[c]{Devices} & Models & Parameters \\
        \midrule
        \makecell*[c]{Hybrid Optical \\System} & \makecell[l]{Inivation DVXplorer\\ \\DAHENG MER2} & \makecell[l]{640$\times$480 1/3.5''\\ FoV: 20°$\times$15°\\ \midrule 1440$\times$1080 \\ FoV: 17°$\times$13° 1/2.9'' \\ Rate: 25Hz }\\
        \midrule
        \makecell*[c]{Linear Rail\\System} & RXP45-2000 & \makecell[l]{Stroke Pitch: 2000mm\\ Max Speed: 1500mm/s \\ Pos. Accuracy: ±0.05mm\\ Maximum Load: 10kg }\\
        \bottomrule 
    \end{tabular}%
    }
    \caption{Setup and parameters of the small-scale TTC testbed.}
    \label{tab:hybird_optical_system}
\vspace{-1.5cm}
\end{table}

%% file: floats/tab_benchmark.tex
\begin{table*}[htb]
\centering
\begin{adjustbox}{max width=\linewidth}
\setlength{\tabcolsep}{3pt}
% \begin{tabular}{cc|cccccccc}
\begin{tabular}{l|l|l|l|l|l|l|l|l|l|l}

\toprule

\multirow{2}*{Method}
& \multicolumn{2}{c}{CCRs1-low} & \multicolumn{2}{c}{CCRs1-medium}& \multicolumn{2}{c}{CCRs1-high} & \multicolumn{2}{c}{CCRs2-low} & \multicolumn{2}{c}{CCRs2-medium} 
\\

\cmidrule{2-11}
            & $e_{\mathrm{TTC} }~(\%)$      &Runtime~(s)
            & $e_{\mathrm{TTC} }~(\%)$      &Runtime~(s)
            & $e_{\mathrm{TTC} }~(\%)$      &Runtime~(s)
            & $e_{\mathrm{TTC} }~(\%)$      &Runtime~(s)
            & $e_{\mathrm{TTC} }~(\%)$      &Runtime~(s)
\\
\midrule
STRTTC\cite{li2024eventaidedtimetocollisionestimationautonomous}    & 7.54              &\uline{0.031}      & 9.64 & \uline{0.026}  & 7.78 & 0.045  & 8.15 & \uline{0.024} & 11.83 & \uline{0.024} 
\\
CMax\cite{gallego2018unifying}                                      &\textbf{2.56}      &3.833              & \textbf{3.44} & 3.701  & \uline{7.39} & 3.228  & \uline{5.02} & 2.32  & \uline{11.20}  &3.449
\\
ETTCM\cite{nunes2023time}                                           &45.11              &  0.081            & 44.94        & 0.081  & 43.03  & 0.300  & 48.83 & 0.350 & 52.84 & 0.098  
\\
FAITH\cite{dinaux2021faith}                                         &15.15             & 0.156                 &  20.63 & 0.155  & 19.51 & 0.337 &25.83  & 0.169   &47.45   & 0.191 
\\
AEB-Tracker\cite{wang2024asynchronous}                              &39.57   & $\mathbf{4.2\times10^{-5}}$  &42.41   &$\mathbf{3.9\times10^{-5}}$  & 39.04 & $\mathbf{1.1\times10^{-5}}$ & 37.09 & $\mathbf{9.4\times10^{-6}}$ & 43.17 & $\mathbf{7.3\times10^{-6}}$

% \textbf{4.2 × 10}
% $\mathbf{4.2\times10^{-5}}$
\\
Image's FoE\cite{stabinger2016monocular}                            &\uline{5.37}      & 0.036        & \uline{5.12} & 0.031  & \textbf{1.86} & \uline{0.041}  & \textbf{3.76}  & 0.031   & \textbf{3.85}   & 0.031  
\\
% \Xhline{1pt}
\bottomrule
\toprule
\multirow{2}*{Method}
& \multicolumn{2}{c}{CCRs2-high} & \multicolumn{2}{c}{CCRm-low} & \multicolumn{2}{c}{CCRm-medium}  & \multicolumn{2}{c}{Slider-750} & \multicolumn{2}{c}{Slider-1000}
\\
\cmidrule{2-11}
            & $e_{\mathrm{TTC} }~(\%)  $    &Runtime~(s)
            & $e_{\mathrm{TTC} }~(\%)  $    &Runtime~(s)
            & $e_{\mathrm{TTC} }~(\%)  $    &Runtime~(s)
            & $e_{\mathrm{TTC} }~(\%)  $    &Runtime~(s)
            & $e_{\mathrm{TTC} }~(\%)  $    &Runtime~(s)
            
\\
\midrule
STRTTC\cite{li2024eventaidedtimetocollisionestimationautonomous}   & 13.68 & \uline{0.028}      & 9.87 & 0.031      & \uline{10.36} & \uline{0.023}      & \uline{8.95} & 0.015      & 12.43 &0.016 
\\            
CMax\cite{gallego2018unifying}  & \uline{11.10} & 3.826 & \uline{9.27} & 3.658& 14.73 & 3.657  & \textbf{4.16} & 0.85 & \textbf{2.74} & 0.93
\\
ETTCM\cite{nunes2023time}    & 49.46 & 0.127 & 71.05 & 0.059 & 58.00  & 0.121  & 18.99 & 0.498 & 15.20 & 0.191
\\
FAITH\cite{dinaux2021faith}   & 49.62 & 0.165 &75.78 & 0.116 &56.26  &  0.11  & 34.45 & 0.186 & 47.94 & 0.233
\\
AEB-Tracker\cite{wang2024asynchronous}  & 42.43 & $\mathbf{6\times10^{-6}}$ & 26.95 & $\mathbf{1\times10^{-5}}$ & 35.99 & $\mathbf{1.1\times10^{-5}}$ & 157.88 & $\mathbf{1\times10^{-5}}$ & 165.56 & $\mathbf{7.8\times10^{-6}}$
\\
Image's FoE\cite{stabinger2016monocular}   & \textbf{2.92} & 0.031 & \textbf{5.60} & \uline{0.027} & \textbf{3.86}  & 0.026 & 11.35 & \uline{0.013} & \uline{7.55} & \uline{0.012}  
\\
\bottomrule

\end{tabular}

\end{adjustbox}
\caption{\textit{Performance of multiple TTC estimation algorithms.} Best and second best results
are \textbf{highlighted} and \uline{underlined}, respectively.} 
% \vspace{-0.5cm}
\label{tab: benchmark}
\end{table*}

%% file: chapters/05_Experiments.tex
\section{TTC Benchmark}
\label{sec: experiments}
With the above dataset and testing platform, we provide a publicly accessible benchmark to the community.
To this end, we evaluate a number of TTC estimation algorithms, including STRTTC~\cite{li2024eventaidedtimetocollisionestimationautonomous}, CMax~\cite{gallego2018unifying}, ETTCM~\cite{nunes2023time}, FAITH~\cite{dinaux2021faith}, AEB-Tracker~\cite{wang2024asynchronous} and Image’s FOE~\cite{stabinger2016monocular}, using our dataset.
All evaluation results are obtained by averaging over multiple runs.
The results of these algorithms are presented in Tab.~\ref{tab: benchmark}.

{STRTTC~\cite{li2024eventaidedtimetocollisionestimationautonomous} demonstrates robust performance under various conditions, attributed to its resilient sampling strategy and efficient initialization scheme.
Among event-based methods, CMax~\cite{gallego2018unifying} achieves the highest accuracy by utilizing all events within the bounding box, thereby enhancing the signal-to-noise ratio. 
However, its nonlinear least-squares optimization adds computational complexity, with processing 3e5 events typically taking around 3 seconds.
ETTCM~\cite{nunes2023time} estimates the TTC and reports computation time per event. 
For easier comparison, we define its computation time as the product of the time per computation and the total number of events processed. 
In our experiments, while it shows high efficiency, it does not scale well with input data generated per unit time.
Using event-based normal flow as input, FAITH~\cite{dinaux2021faith} employs a RANSAC-based algorithm to determine the FOE, demonstrating real-time performance.
However, substituting normal flow for optical flow leads to a decrease in the accuracy of TTC estimation.

AEB-Tracker~\cite{wang2024asynchronous} achieves exceptionally high computational efficiency, reaching up to 10 kHz. 
However, under conditions with significant background noise, the algorithm requires a low-frequency detector to identify regions (\textit{e.g.}, two taillights) capable of generating event blobs. 
Furthermore, it requires that the tracked event blobs adhere to a spatio-temporal Gaussian likelihood model~\cite{wang2024asynchronous}; otherwise, the accuracy of the estimated TTC may decrease.
Image’s FoE~\cite{stabinger2016monocular} takes RGB images from a camera running at 20 Hz as input. 
It achieves high accuracy and efficiency across all sequences. 
Nevertheless, its performance is constrained by the limitations of standard cameras (\textit{e.g.}, data latency, motion blur).

%% file: chapters/06_Conclusion.tex
\section{Conclusion}
\label{sec:conclusion}
We present an event camera dataset, called EvTTC, for the task of time-to-collision estimation in autonomous driving.
It consists of data captured using frame-based and event-based cameras, covering various scenarios of potential collision in daily driving.
High-precision external reference signals are also provided for ground truth, and a benchmark of event-based TTC is established.
Besides, a small-scale TTC testbed for experimental validation and data augmentation is presented.
We wish the release of EvTTC can facilitate the development of forward collision warning techniques using event cameras.

\section*{Acknowledgment}
We thank Javier Hidalgo-Carri{\'o} and Davide Scaramuzza for releasing the design of the Beamsplitter~\cite{hidalgo2022event}, based on which we build our FCW system. 
We thank Mr. Yanggang Sheng and Mr. Junkai Niu for the help in the data collection. 
We also thank Mr. Sheng Zhong for proofreading.

%% file: chapters/Appendix.tex
\onecolumn
\section*{Supplementary Material}
% \vspace{-3.5cm}
The figure below demonstrates detailed test results for the TTC estimation methods described in Sec.~\ref{sec: experiments}, evaluated on our dataset.
% \vspace{-1cm}
\input{floats/fig_results}
\input{floats/fig_results1}
\twocolumn

%% file: floats/fig_results.tex
\begin{figure}[H]
\centering
\resizebox{0.98\textwidth}{!}{ 
    \begin{minipage}{\textwidth}
        % Row 1
        \subfloat[CCRs1-low]{%
            \includegraphics[width=0.45\textwidth]{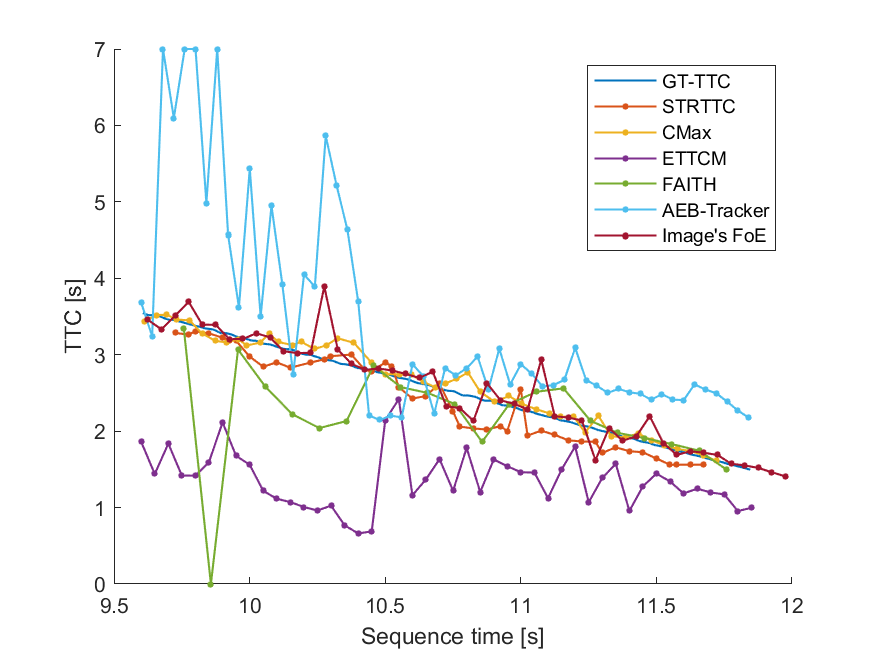}%
            \label{fig:CCRs1-low}
        }
        \hspace{0.05\textwidth}
        \subfloat[CCRs1-medium]{%
            \includegraphics[width=0.45\textwidth]{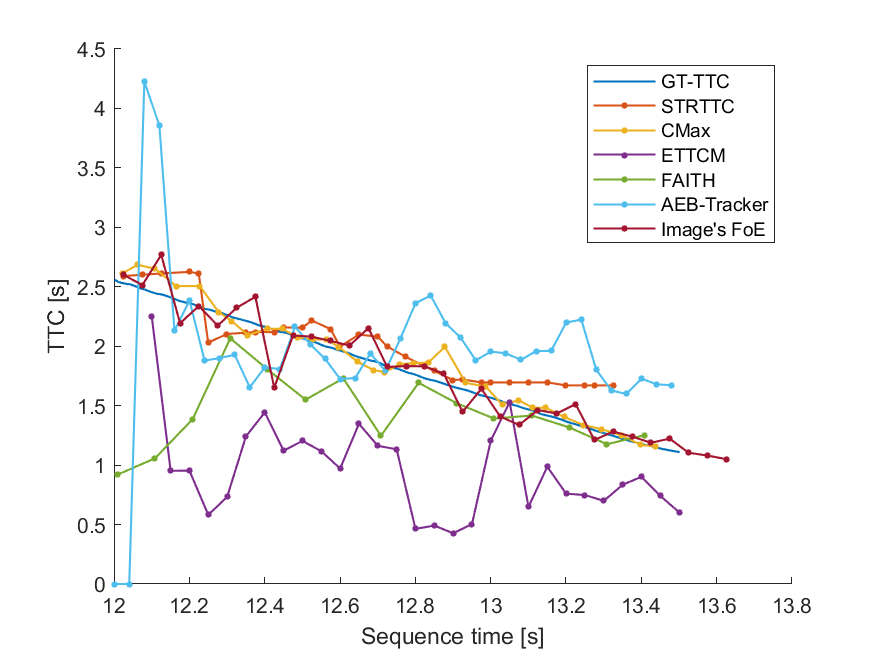}%
            \label{fig:CCRs1-medium}
        } \\[0.3cm]

        % Row 2
        \subfloat[CCRs1-high]{%
            \includegraphics[width=0.45\textwidth]{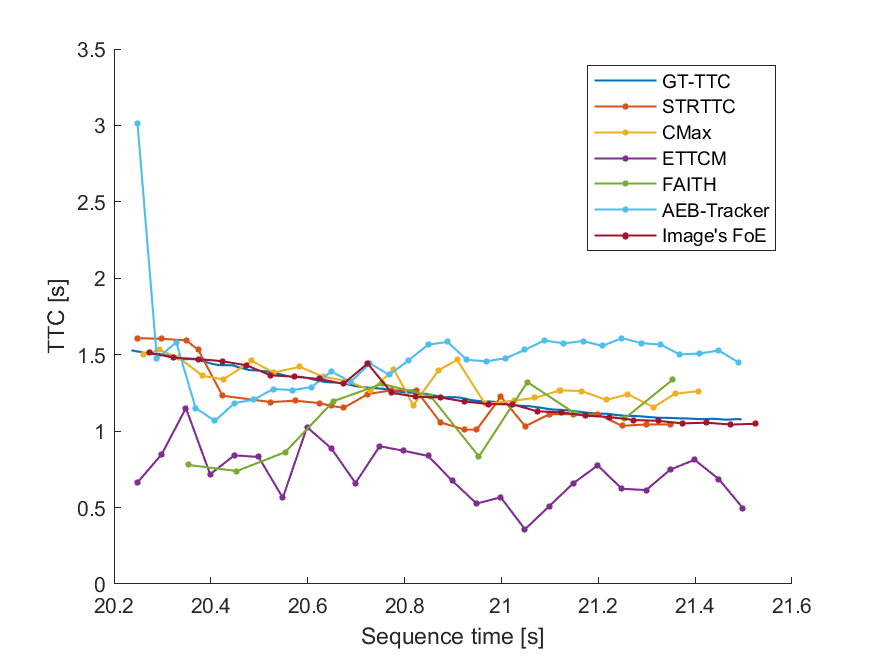}%
            \label{fig:CCRs1-high}
        }
        \hspace{0.05\textwidth}
        \subfloat[CCRs2-low]{%
            \includegraphics[width=0.45\textwidth]{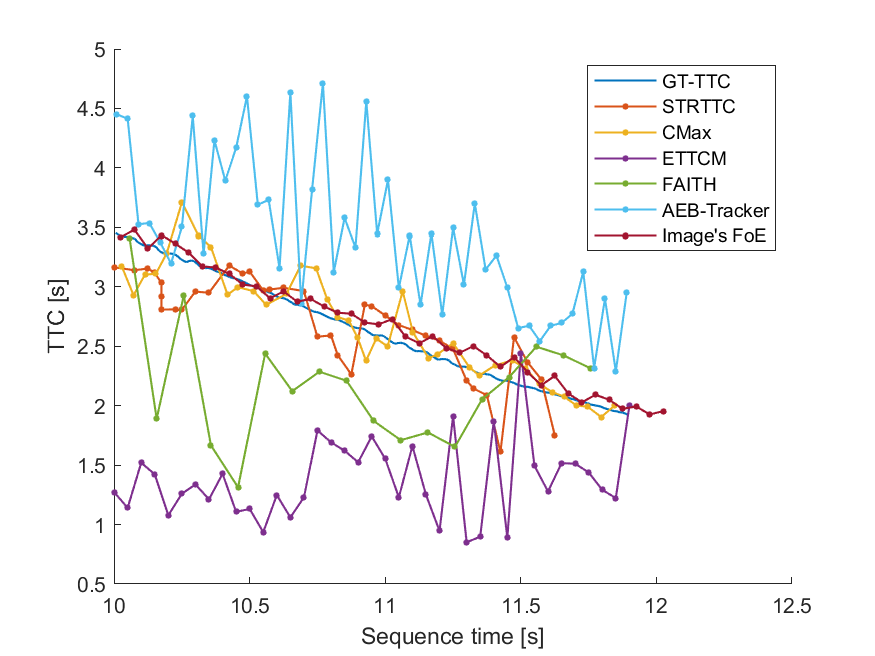}%
            \label{fig:CCRs2-low}
        } \\[0.3cm]

        % Row 3
        \subfloat[CCRs2-medium]{%
            \includegraphics[width=0.45\textwidth]{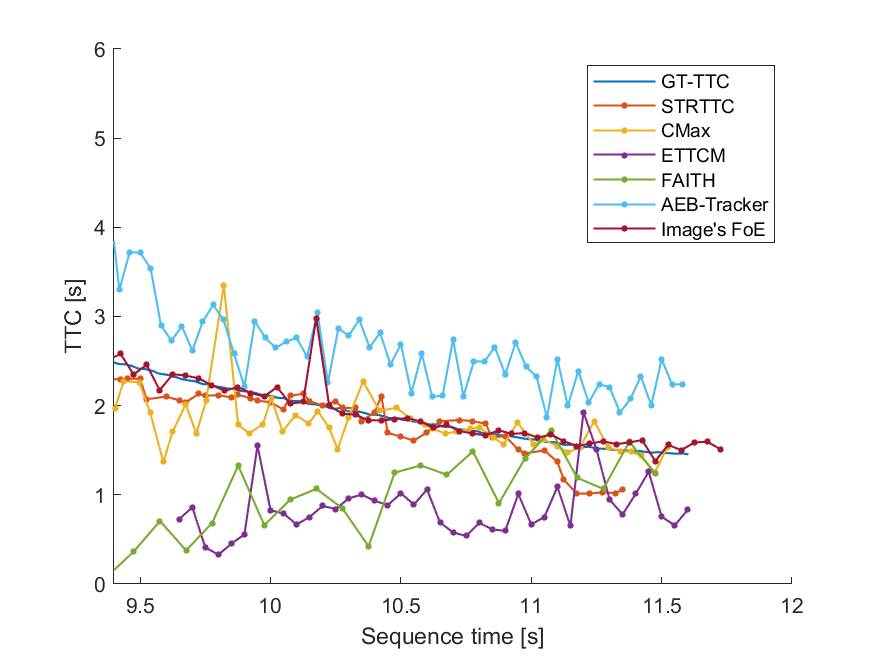}%
            \label{fig:CCRs2-medium}
        }
        \hspace{0.05\textwidth}
        \subfloat[CCRs2-high]{%
            \includegraphics[width=0.45\textwidth]{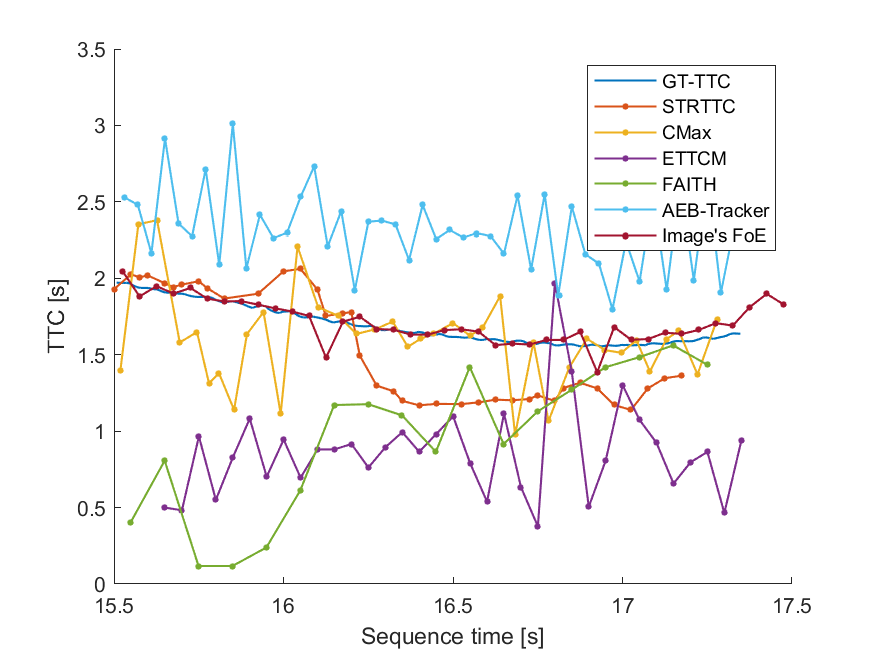}%
            \label{fig:CCRs2-high}
        } \\[0.3cm]

    \end{minipage}
}
\label{fig:results}
\end{figure}

%% file: floats/fig_results1.tex
\begin{figure}[H]
\centering
\resizebox{0.98\textwidth}{!}{
    \begin{minipage}{\textwidth}
        % Row 4
        \subfloat[CCRm-low]{%
            \includegraphics[width=0.45\textwidth]{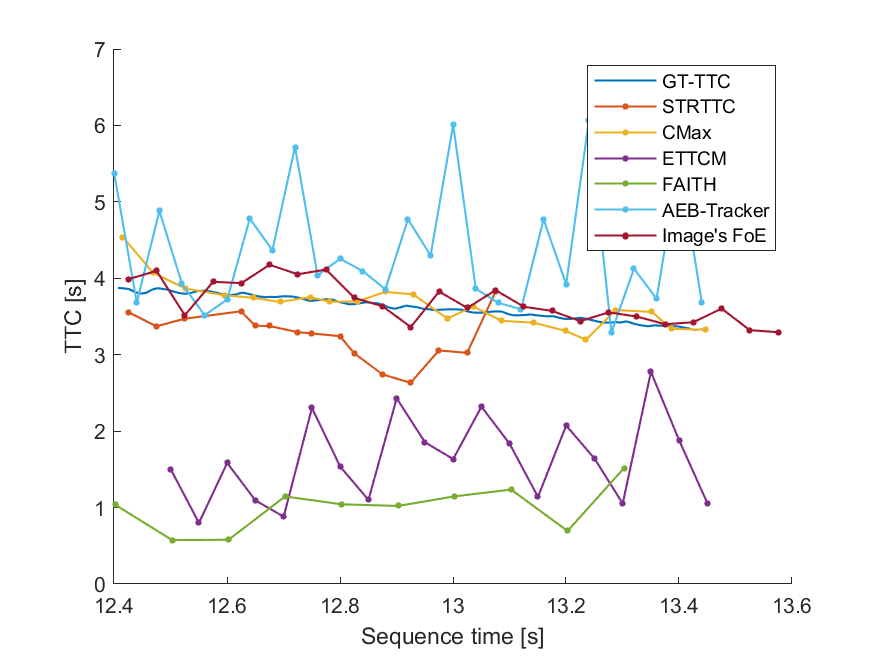}%
            \label{fig:CCRm-low}
        }
        \hspace{0.05\textwidth}
        \subfloat[CCRm-medium]{%
            \includegraphics[width=0.45\textwidth]{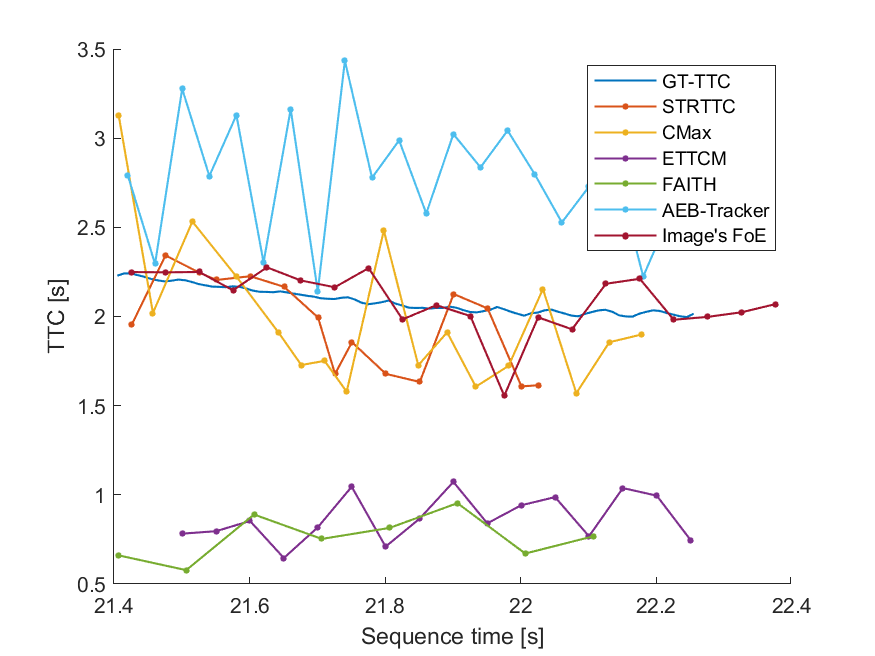}%
            \label{fig:CCRm-medium}
        } \\[0.3cm]
        % Row 5
        \subfloat[slider-750]{%
            \includegraphics[width=0.45\textwidth]{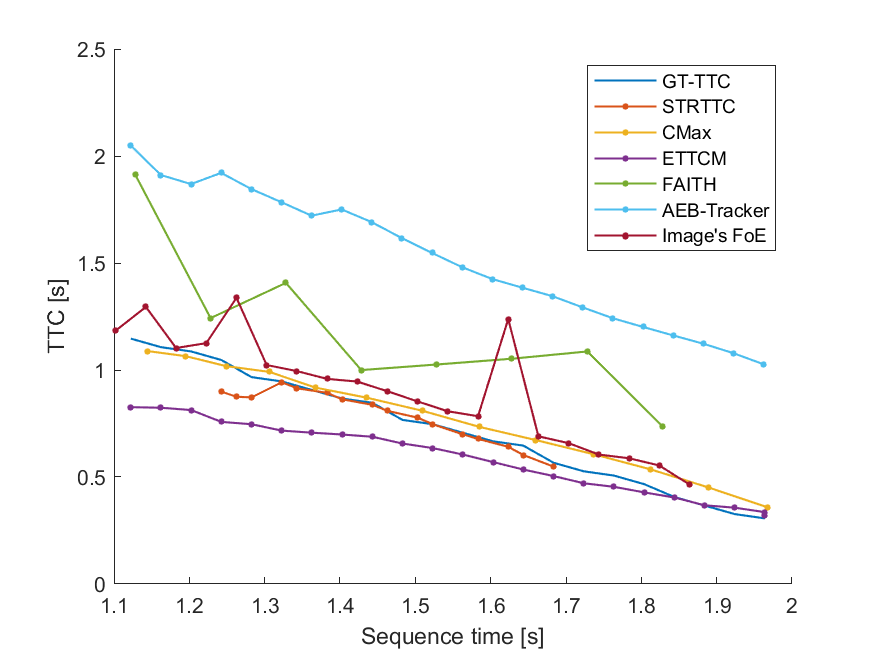}%
            \label{fig:slider-750}
        }
        \hspace{0.05\textwidth}
        \subfloat[slider-1000]{%
            \includegraphics[width=0.45\textwidth]{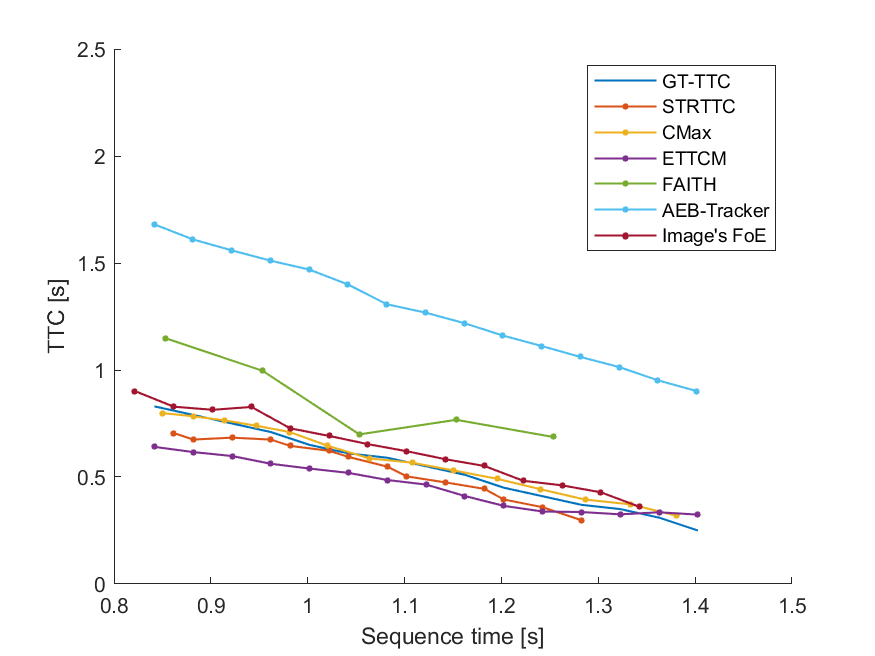}%
            \label{fig:slider-1000}
        } \\[0.3cm]
    \end{minipage}
}
\clearpage 
\label{fig:results}
\end{figure}

%% file: main.bbl
\begin{thebibliography}{10}
\providecommand{\url}[1]{#1}
\csname url@rmstyle\endcsname
\providecommand{\newblock}{\relax}
\providecommand{\bibinfo}[2]{#2}
\providecommand\BIBentrySTDinterwordspacing{\spaceskip=0pt\relax}
\providecommand\BIBentryALTinterwordstretchfactor{4}
\providecommand\BIBentryALTinterwordspacing{\spaceskip=\fontdimen2\font plus
\BIBentryALTinterwordstretchfactor\fontdimen3\font minus \fontdimen4\font\relax}
\providecommand\BIBforeignlanguage[2]{{%
\expandafter\ifx\csname l@#1\endcsname\relax
\typeout{** WARNING: IEEEtran.bst: No hyphenation pattern has been}%
\typeout{** loaded for the language `#1'. Using the pattern for}%
\typeout{** the default language instead.}%
\else
\language=\csname l@#1\endcsname
\fi
#2}}

\bibitem{palazzi2018predicting}
A.~Palazzi, D.~Abati, F.~Solera, R.~Cucchiara, \emph{et~al.}, ``Predicting the driver's focus of attention: the dr (eye) ve project,'' \emph{IEEE transactions on pattern analysis and machine intelligence}, vol.~41, no.~7, pp. 1720--1733, 2018.

\bibitem{fang2019dada}
J.~Fang, D.~Yan, J.~Qiao, J.~Xue, H.~Wang, and S.~Li, ``Dada-2000: Can driving accident be predicted by driver attentionƒ analyzed by a benchmark,'' in \emph{2019 IEEE Intelligent Transportation Systems Conference (ITSC)}.\hskip 1em plus 0.5em minus 0.4em\relax IEEE, 2019, pp. 4303--4309.

\bibitem{kim2019crash}
H.~Kim, K.~Lee, G.~Hwang, and C.~Suh, ``Crash to not crash: Learn to identify dangerous vehicles using a simulator,'' in \emph{Proceedings of the AAAI Conference on Artificial Intelligence}, vol.~33, no.~01, 2019, pp. 978--985.

\bibitem{shi2023tsttc}
Y.~Shi, Z.~Huang, Y.~Yan, N.~Wang, and X.~Guo, ``Tsttc: A large-scale dataset for time-to-contact estimation in driving scenarios,'' \emph{arXiv preprint arXiv:2309.01539}, 2023.

\bibitem{zhu2018multivehicle}
A.~Z. Zhu, D.~Thakur, T.~{\"O}zaslan, B.~Pfrommer, V.~Kumar, and K.~Daniilidis, ``The multivehicle stereo event camera dataset: An event camera dataset for 3d perception,'' \emph{IEEE Robotics and Automation Letters}, vol.~3, no.~3, pp. 2032--2039, 2018.

\bibitem{Gehrig21ral}
M.~Gehrig, W.~Aarents, D.~Gehrig, and D.~Scaramuzza, ``Dsec: A stereo event camera dataset for driving scenarios,'' \emph{IEEE Robotics and Automation Letters}, 2021.

\bibitem{lee2022vivid++}
A.~J. Lee, Y.~Cho, Y.-s. Shin, A.~Kim, and H.~Myung, ``Vivid++: Vision for visibility dataset,'' \emph{IEEE Robotics and Automation Letters}, vol.~7, no.~3, pp. 6282--6289, 2022.

\bibitem{chaney2023m3ed}
K.~Chaney, F.~Cladera, Z.~Wang, A.~Bisulco, M.~A. Hsieh, C.~Korpela, V.~Kumar, C.~J. Taylor, and K.~Daniilidis, ``M3ed: Multi-robot, multi-sensor, multi-environment event dataset,'' in \emph{Proceedings of the IEEE/CVF Conference on Computer Vision and Pattern Recognition}, 2023, pp. 4015--4022.

\bibitem{cicchino2017effectiveness}
J.~B. Cicchino, ``Effectiveness of forward collision warning and autonomous emergency braking systems in reducing front-to-rear crash rates,'' \emph{Accident Analysis \& Prevention}, vol.~99, pp. 142--152, 2017.

\bibitem{fildes2015effectiveness}
B.~Fildes, M.~Keall, N.~Bos, A.~Lie, Y.~Page, C.~Pastor, L.~Pennisi, M.~Rizzi, P.~Thomas, and C.~Tingvall, ``Effectiveness of low speed autonomous emergency braking in real-world rear-end crashes,'' \emph{Accident Analysis \& Prevention}, vol.~81, pp. 24--29, 2015.

\bibitem{isaksson2015evaluation}
I.~Isaksson-Hellman and M.~Lindman, ``Evaluation of rear-end collision avoidance technologies based on real world crash data,'' \emph{Proceedings of the Future Active Safety Technology Towards zero traffic accidents (FASTzero), Gothenburg, Sweden}, pp. 9--11, 2015.

\bibitem{rizzi2014injury}
M.~Rizzi, A.~Kullgren, and C.~Tingvall, ``Injury crash reduction of low-speed autonomous emergency braking (aeb) on passenger cars,'' in \emph{International Research Council on the Biomechanics of Injury (IRCOBI 2014)}.\hskip 1em plus 0.5em minus 0.4em\relax International Research Council on the Biomechanics of Injury, 2014, pp. 656--665.

\bibitem{gawande2022autonomous}
M.~Gawande, P.~Rajalakshmi, \emph{et~al.}, ``Autonomous emergency breaking (aeb) evaluation for indian traffic scenarios using gps and lidar data,'' in \emph{2022 IEEE IAS Global Conference on Emerging Technologies (GlobConET)}.\hskip 1em plus 0.5em minus 0.4em\relax IEEE, 2022, pp. 655--660.

\bibitem{bosnak2017efficient}
M.~Bosnak and I.~Skrjanc, ``Efficient time-to-collision estimation for a braking supervision system with lidar,'' in \emph{2017 3rd IEEE International Conference on Cybernetics (CYBCONF)}.\hskip 1em plus 0.5em minus 0.4em\relax IEEE, 2017, pp. 1--6.

\bibitem{kotur2021camera}
M.~Kotur, N.~Luki{\'c}, M.~Kruni{\'c}, and {\v{Z}}.~Luka{\v{c}}, ``Camera and lidar sensor fusion for 3d object tracking in a collision avoidance system,'' in \emph{2021 Zooming Innovation in Consumer Technologies Conference (ZINC)}.\hskip 1em plus 0.5em minus 0.4em\relax IEEE, 2021, pp. 198--202.

\bibitem{venkatesha2023detection}
K.~Venkatesha, A.~Vengadarajan, and U.~Singh, ``Detection mechanism for vehicle collision avoidance using mmwave radar,'' in \emph{2023 IEEE International Conference on Electronics, Computing and Communication Technologies (CONECCT)}.\hskip 1em plus 0.5em minus 0.4em\relax IEEE, 2023, pp. 1--5.

\bibitem{dagan2004forward}
E.~Dagan, O.~Mano, G.~P. Stein, and A.~Shashua, ``Forward collision warning with a single camera,'' in \emph{IEEE Intelligent Vehicles Symposium, 2004}.\hskip 1em plus 0.5em minus 0.4em\relax IEEE, 2004, pp. 37--42.

\bibitem{meyer1992estimation}
F.~Meyer and P.~Bouthemy, ``Estimation of time-to-collision maps from first order motion models and normal flows,'' in \emph{1992 11th IAPR International Conference on Pattern Recognition}, vol.~1.\hskip 1em plus 0.5em minus 0.4em\relax IEEE Computer Society, 1992, pp. 78--82.

\bibitem{negre2008real}
A.~Negre, C.~Braillon, J.~L. Crowley, and C.~Laugier, ``Real-time time-to-collision from variation of intrinsic scale,'' in \emph{Experimental Robotics: The 10th International Symposium on Experimental Robotics}.\hskip 1em plus 0.5em minus 0.4em\relax Springer, 2008, pp. 75--84.

\bibitem{stabinger2016monocular}
S.~Stabinger, A.~Rodriguez-Sanchez, and J.~Piater, ``Monocular obstacle avoidance for blind people using probabilistic focus of expansion estimation,'' in \emph{2016 IEEE Winter Conference on Applications of Computer Vision (WACV)}.\hskip 1em plus 0.5em minus 0.4em\relax IEEE, 2016, pp. 1--9.

\bibitem{falanga2020dynamic}
D.~Falanga, K.~Kleber, and D.~Scaramuzza, ``Dynamic obstacle avoidance for quadrotors with event cameras,'' \emph{Science Robotics}, vol.~5, no.~40, p. eaaz9712, 2020.

\bibitem{rebecq2017real}
H.~Rebecq, T.~Horstschaefer, and D.~Scaramuzza, ``Real-time visual-inertial odometry for event cameras using keyframe-based nonlinear optimization,'' 2017.

\bibitem{lu2023event}
X.~Lu, Y.~Zhou, and S.~Shen, ``Event-based visual inertial velometer,'' \emph{Robotics: Science and Systems (RSS)}, 2024.

\bibitem{gehrig2024low}
D.~Gehrig and D.~Scaramuzza, ``Low-latency automotive vision with event cameras,'' \emph{Nature}, vol. 629, no. 8014, pp. 1034--1040, 2024.

\bibitem{clady2014asynchronous}
X.~Clady, C.~Clercq, S.-H. Ieng, F.~Houseini, M.~Randazzo, L.~Natale, C.~Bartolozzi, and R.~Benosman, ``Asynchronous visual event-based time-to-contact,'' \emph{Frontiers in neuroscience}, vol.~8, p.~9, 2014.

\bibitem{dupeyroux2021neuromorphic}
J.~Dupeyroux, J.~J. Hagenaars, F.~Paredes-Vall{\'e}s, and G.~C. de~Croon, ``Neuromorphic control for optic-flow-based landing of mavs using the loihi processor,'' in \emph{2021 IEEE International Conference on Robotics and Automation (ICRA)}.\hskip 1em plus 0.5em minus 0.4em\relax IEEE, 2021, pp. 96--102.

\bibitem{dinaux2021faith}
R.~Dinaux, N.~Wessendorp, J.~Dupeyroux, and G.~C. De~Croon, ``Faith: Fast iterative half-plane focus of expansion estimation using optic flow,'' \emph{IEEE Robotics and Automation Letters}, vol.~6, no.~4, pp. 7627--7634, 2021.

\bibitem{gallego2018unifying}
G.~Gallego, H.~Rebecq, and D.~Scaramuzza, ``A unifying contrast maximization framework for event cameras, with applications to motion, depth, and optical flow estimation,'' in \emph{Proceedings of the IEEE conference on computer vision and pattern recognition}, 2018, pp. 3867--3876.

\bibitem{shiba2023fast}
S.~Shiba, Y.~Aoki, and G.~Gallego, ``A fast geometric regularizer to mitigate event collapse in the contrast maximization framework,'' \emph{Advanced Intelligent Systems}, vol.~5, no.~3, p. 2200251, 2023.

\bibitem{nunes2023time}
U.~M. Nunes, L.~U. Perrinet, and S.-H. Ieng, ``Time-to-contact map by joint estimation of up-to-scale inverse depth and global motion using a single event camera,'' in \emph{Proceedings of the IEEE/CVF International Conference on Computer Vision}, 2023, pp. 23\,653--23\,663.

\bibitem{nunes2021robust}
U.~M. Nunes and Y.~Demiris, ``Robust event-based vision model estimation by dispersion minimisation,'' \emph{IEEE Transactions on Pattern Analysis and Machine Intelligence}, vol.~44, no.~12, pp. 9561--9573, 2021.

\bibitem{li2024eventaidedtimetocollisionestimationautonomous}
J.~Li, B.~Liao, X.~Lu, P.~Liu, S.~Shen, and Y.~Zhou, ``Event-aided time-to-collision estimation for autonomous driving,'' in \emph{European Conference on Computer Vision}.\hskip 1em plus 0.5em minus 0.4em\relax Springer, 2025, pp. 57--73.

\bibitem{wang2024asynchronous}
Z.~Wang, T.~Molloy, P.~van Goor, and R.~Mahony, ``Asynchronous blob tracker for event cameras,'' \emph{IEEE Transactions on Robotics}, 2024.

\bibitem{ieee1588-2019}
{IEEE Std 1588-2019 (Revision of IEEE Std 1588-2008)}, ``{IEEE Standard for a Precision Clock Synchronization Protocol for Networked Measurement and Control Systems},'' pp. 1--499, 2020.

\bibitem{ieee8021as2020}
{IEEE Std 802.1AS-2020 (Revision of IEEE Std 802.1AS-2011)}, ``{IEEE Standard for Local and Metropolitan Area Networks--Timing and Synchronization for Time-Sensitive Applications},'' pp. 1--421, 2020.

\bibitem{furgale2013unified}
P.~Furgale, J.~Rehder, and R.~Siegwart, ``Unified temporal and spatial calibration for multi-sensor systems,'' in \emph{2013 IEEE/RSJ International Conference on Intelligent Robots and Systems}.\hskip 1em plus 0.5em minus 0.4em\relax IEEE, 2013, pp. 1280--1286.

\bibitem{opencalib}
G.~Yan, Z.~Liu, C.~Wang, C.~Shi, P.~Wei, X.~Cai, T.~Ma, Z.~Liu, Z.~Zhong, Y.~Liu, M.~Zhao, Z.~Ma, and Y.~Li, ``Opencalib: A multi-sensor calibration toolbox for autonomous driving,'' \emph{arXiv preprint arXiv:2205.14087}, 2022.

\bibitem{euroNCAP2023AEB}
\BIBentryALTinterwordspacing
E.~NCAP, ``Euro ncap aeb c2c test protocol - v4.3.1,'' 2024. [Online]. Available: \url{https://www.euroncap.com/media/80155/euro-ncap-aeb-c2c-test-protocol-v431.pdf}
\BIBentrySTDinterwordspacing

\bibitem{xu2022fast}
W.~Xu, Y.~Cai, D.~He, J.~Lin, and F.~Zhang, ``Fast-lio2: Fast direct lidar-inertial odometry,'' \emph{IEEE Transactions on Robotics}, vol.~38, no.~4, pp. 2053--2073, 2022.

\bibitem{katz2007direct}
S.~Katz, A.~Tal, and R.~Basri, ``Direct visibility of point sets,'' in \emph{ACM SIGGRAPH 2007 papers}, 2007, pp. 24--es.

\bibitem{hidalgo2022event}
J.~Hidalgo-Carri{\'o}, G.~Gallego, and D.~Scaramuzza, ``Event-aided direct sparse odometry,'' in \emph{Proceedings of the IEEE/CVF Conference on Computer Vision and Pattern Recognition}, 2022, pp. 5781--5790.

\end{thebibliography}
